\documentclass[sigconf]{acmart}

\AtBeginDocument{%
  }

\usepackage{threeparttable}
\usepackage{booktabs}
\usepackage{colortbl}
\usepackage[table]{xcolor}
\usepackage{pifont}

\definecolor{oursblue}{RGB}{220, 235, 255}
\usepackage[table]{xcolor}
\usepackage{colortbl}
\usepackage{enumitem}
\usepackage{multirow}
\usepackage{algorithm}
\usepackage{algorithmic}
\usepackage{graphicx}
\usepackage{url}
\usepackage{dblfloatfix}
\usepackage{tcolorbox}
\tcbuselibrary{most}
\newtcolorbox{promptbox}[1][]{colback=gray!5,colframe=gray!75,fonttitle=\bfseries,title=Prompt,#1}

\acmConference[MM'26]{the 34th ACM International Conference on Multimedia}{November 10-14, 2026}{Rio de Janeiro, Brazil}

\begin{document}

\title{EgoProceVQA: A Novel Egocentric Procedural Understanding Task with Self-Skill-Exploration Agent}

\author{Junlong Li$^*$}
%\authornote{Equally Contribution}
\affiliation{%
  \institution{The Hong Kong Polytechnic University}
  \city{Hong Kong}
  \country{China}
}
\email{junlong.li@connect.polyu.hk}

\author{Junxi Li$^*$}
%\authornote{Equally Contribution}
\affiliation{%
  \institution{The Hong Kong Polytechnic University}
  \city{Hong Kong}
  \country{China}
}
\email{lee2333.li@connect.polyu.hk}
\thanks{$*$ Equal contribution, † Corresponding author}
%\correspondingauthor
\author{Yuxiang Yang}
%\authornote{Equally Contribution}
\affiliation{%
  \institution{The Hong Kong Polytechnic University}
  \city{Hong Kong}
  \country{China}
}

\author{Wenbin Zou}
%\authornote{Equally Contribution}
\affiliation{%
  \institution{The Hong Kong Polytechnic University}
  \city{Hong Kong}
  \country{China}\\
  \institution{South China University of Technology}
  \city{Guang Zhou}
  \country{China}
}
\author{Lap-Pui Chau}
%\authornote{Equally Contribution}
\affiliation{%
  \institution{The Hong Kong Polytechnic University}
  \city{Hong Kong}
  \country{China}
}
\author{Yi Wang$^\dagger$}
\affiliation{%
  \institution{The Hong Kong Polytechnic University}
  \city{Hong Kong}
  \country{China}
}
\email{yi-eie.wang@polyu.edu.hk}

\begin{abstract}

    Most daily activities are inherently procedural. However, existing evaluations for egocentric video understanding seldom address procedural understanding and largely overlook complex key-step-level reasoning under the widely used video question answering (VQA) paradigm for MLLMs. Such capabilities are crucial for building procedural AI assistants deployable on wearable devices. To bridge this gap, we introduce the Egocentric Procedural Understanding VQA task (EgoProceVQA), which systematically evaluates egocentric procedural reasoning abilities of current MLLMs and agents through six types of key-step-centric questions. Furthermore, we develop EgoProceGen, a data generation platform that efficiently constructs QA data tailored to different question types. Based on this platform, we build a benchmark with 3,600 questions, four common procedural scenarios, and 31 everyday procedural tasks. Evaluations on EgoProceVQA show that existing MLLMs and agents still have substantial room for improvement in procedural understanding. Therefore, we further propose EgoProceAgent, a self-skill-exploration agentic framework. We design a generic tool library for procedural understanding and a standardized sub-skill library shared across tools and models, enabling self-exploration without ground-truth supervision. By exploring how to compose and select sub-skills, the agent discovers effective skill strategies for diverse problems, and attains state-of-the-art performance among open-source models on multiple tasks. Together, our benchmark, generation platform, and agentic framework establish a unified foundation for EgoProceVQA. Project page: \url{https://z1oong.github.io/EgoProceVQA/}.

\end{abstract}

\keywords{Egocentric vision, procedural understanding, agent skill, self-evolution}

\maketitle
\begin{figure}
    \centering
    \includegraphics[width=1\linewidth]{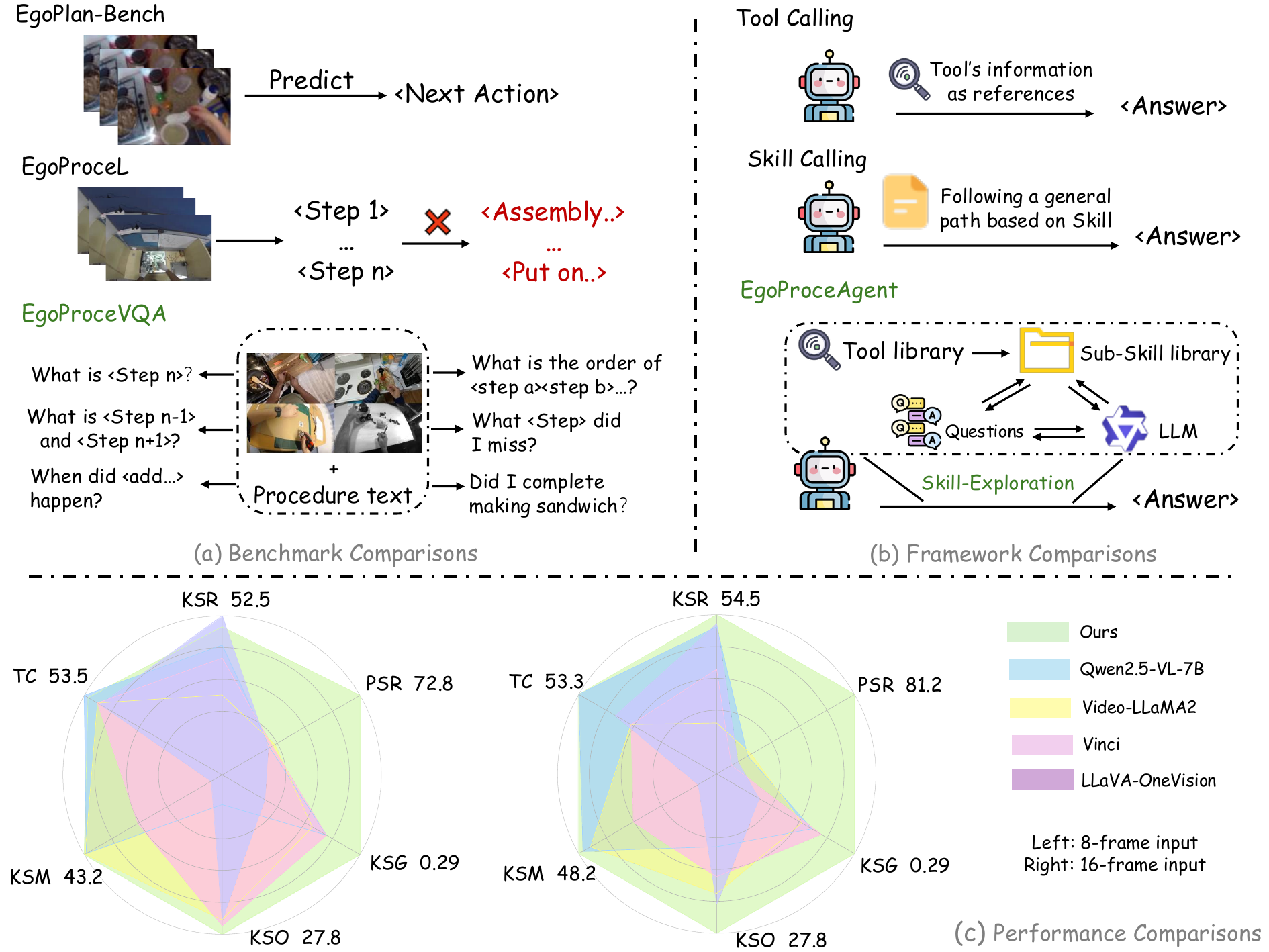}
\vspace{-15pt}
    \caption{Overview of EgoProceVQA. (a) demonstrates that we uniquely introduce a benchmark for key-step-level VQA across six question types. (b) shows that EgoProceAgent autonomously explores and constructs its own skills, and subsequently leverages these skills for tool invocation and reasoning. (c) shows our outstanding performance.}
    \label{fig1}
    \vspace{-18pt}
\end{figure}

\section{Introduction}

Egocentric video understanding has become an important testbed for multimodal perception and reasoning, driven by the rapid development of wearable devices \cite{plizzari2024outlook}, embodied assistants \cite{li2025building}, and robots \cite{zheng2026egoscale} that perceive the world from a first-person perspective. Existing egocentric benchmarks have substantially advanced progress on generic understanding tasks such as object recognition, action recognition, and event understanding \cite{fan2019egovqa}. However, many real-world settings in which AI assistance is most needed, such as cooking, assembly, maintenance, and daily task guidance, are inherently procedural: they require the agent not only to recognize what is happening, but also to understand how a task unfolds through a sequence of semantically meaningful and temporally organized key steps. Despite its practical importance, this capability remains insufficiently studied in current egocentric evaluation.

A central limitation of existing benchmarks is that they do not systematically assess key-step-centric procedural understanding. Some prior works consider related abilities such as hierarchical reasoning or future-step prediction \cite{cheng2024videogothink,chen2023egoplan}, but they do not explicitly evaluate whether models can identify, verify, and reason about the critical steps required to complete a procedure. Other datasets focus on task assistance settings with open-ended interaction formats \cite{wong2022assistq,wang2023holoassist}, which are valuable for application development but do not provide a unified, objective standard for fine-grained comparison across models. Meanwhile, datasets closer to procedural learning often emphasize visual key-step sequence recognition alone \cite{bansal2022myview}, with limited text-vision alignment and insufficient support for evaluating modern Multimodal Large Language Models (MLLMs) and QA-based agents. As a result, current benchmarks leave an important gap between generic egocentric understanding and the procedural reasoning abilities required by real-world AI assistants.

\begin{table*}[t]
\vspace{-6pt}
\centering
\caption{%
  Comparison of egocentric video benchmarks.
  \textbf{Step Ann.}~=~step-level annotation;
  \textbf{Temp.\ Order}~=~temporal step-order reasoning;
  \textbf{Task Comp.}~=~task completeness understanding;
  \textbf{Crit.\ Step}~=~critical step identification;
  \textbf{Multi-task}~=~multiple distinct reasoning task types.
  V~=~video; I~=~image; T~=~text; E~=~egocentric.
  Bold \checkmark\ marks capabilities unique to EgoProceVQA.%
}
\label{tab1}
\vspace{-5pt}
\scalebox{0.9}{
\renewcommand{\arraystretch}{1.45}
\resizebox{\textwidth}{!}{%
\large
\begin{tabular}{l l r c c  c  c c c c  l}
\toprule
\multirow{2}{*}{\textbf{Benchmark}}
  & \multirow{2}{*}{\textbf{Year}}
  & \multirow{2}{*}{\textbf{Scale}}
  & \multirow{2}{*}{\textbf{Mod.}}
  & \multirow{2}{*}{\textbf{View}}
  & \multirow{2}{*}{\textbf{Step Ann.}}
  & \multicolumn{4}{c}{\textbf{Reasoning Capability}}
  & \multirow{2}{*}{\textbf{Primary Task}} \\
\cmidrule(lr){7-10}
  & & & & &
  & \textbf{Temp.\ Order}
  & \textbf{Task Comp.}
  & \textbf{Crit.\ Step}
  & \textbf{Multi-task}
  & \\
\midrule
AssistQ~\cite{wong2022assistq}
  & 2022 & 100 vid, 531 QA    & V+T & E
  & $\times$    & $\times$ & $\times$ & $\times$ & $\times$
  & Instruction-following VQA \\
EgoTaskQA~\cite{jia2022egotaskqa}
  & 2022 & 2K vid, 40K QA     & V+T & E
  & \checkmark  & $\times$ & $\times$ & $\times$ & $\times$
  & Goal \& State QA \\
EgoSchema~\cite{mangalam2023egoschema}
  & 2023 & 250 hrs, 5063 QA   & V+T & E
  & $\times$    & $\times$ & $\times$ & $\times$ & $\times$
  & Long-form VQA \\
EgoPlan-Bench~\cite{chen2023egoplan}
  & 2023 & ---, 4939 QA       & V+T & E
  & \checkmark  & $\times$ & $\times$ & $\times$ & $\times$
  & Procedural Planning \\
Ego4D Goal-Step~\cite{nagarajan2023ego4d_goalstep}
  & 2023 & 430 hrs, 48K seg   & V+T & E
  & \checkmark  & $\times$ & $\times$ & $\times$ & $\times$
  & Step Prediction \\
EgoThink~\cite{cheng2024egothink}
  & 2024 & 750 QA, 12 tasks   & I+T & E
  & $\times$    & $\times$ & $\times$ & $\times$ & \checkmark
  & Egocentric General Eval \\
OpenEQA~\cite{majumdar2024openeqa}
  & 2024 & 180+ env, 1636 QA  & V+T & E
  & $\times$    & $\times$ & $\times$ & $\times$ & \checkmark
  & Embodied QA \\
MM-Ego~\cite{mmego2024}
  & 2024 & 7M QA              & V+T & E
  & $\times$    & $\times$ & $\times$ & $\times$ & \checkmark
  & Egocentric General Eval \\
VidEgoThink~\cite{cheng2024videogothink}
  & 2024 & 195 vid, 600 QA    & V+T & E
  & $\times$    & $\times$ & $\times$ & $\times$ & \checkmark
  & Egocentric General Eval \\
EgoTextVQA~\cite{zhou2025egotextvqa}
  & 2025 & 1.5K vid, 7K QA    & V+T & E
  & $\times$    & $\times$ & $\times$ & $\times$ & $\times$
  & Scene-text VQA \\
ProMQA~\cite{hasegawa2025promqa}
  & 2025 & 384 vid, 401 QA    & V+T & E
  & \checkmark  & $\times$ & $\times$ & $\times$ & $\times$
  & LLM-judge open-ended QA \\
\midrule
\rowcolor{gray!12}
\textbf{EgoProceVQA~(Ours)}
  & \textbf{2026}& \textbf{3600 QA (6$\boldsymbol{\times}$600)}
  & \textbf{V+T}
  & \textbf{E}
  & \checkmark
  & $\boldsymbol{\checkmark}$
  & $\boldsymbol{\checkmark}$
  & $\boldsymbol{\checkmark}$
  & ${\checkmark}$
  & \textbf{Key-step-level Procedural VQA} \\
\bottomrule
\end{tabular}%
}
}
\vspace{1ex}

\vspace{-10pt}
\end{table*}

To address this gap, we introduce \textbf{EgoProceVQA}, a new task for systematically evaluating egocentric procedural understanding in MLLMs and agents. Concretely, EgoProceVQA is built around a question-answering (QA) formulation that decomposes procedural understanding into six progressively structured, key-step-centric dimensions. These dimensions move beyond generic perception and probe whether a model can reason about critical procedural units under increasing difficulty, including abilities such as key-step recognition, cross-modal key-step order verification, and temporal grounding. By organizing evaluation around interpretable procedural subtasks rather than a single aggregate task, EgoProceVQA enables more fine-grained diagnosis of model capabilities and failure modes. In this sense, EgoProceVQA is designed not only to measure overall performance, but also to reveal which aspects of procedural understanding remain challenging for current systems.

To support the scale and structural demands of this new evaluation task, we develop \textbf{EgoProceGen}, an automatic data generation platform tailored to procedural QA creation. This platform enables scalable and controllable benchmark construction by supporting the structured generation of multiple procedural subtasks, allowing evaluation protocols to be flexibly instantiated across diverse activity scenarios. Using this platform, we construct a benchmark containing 3,600 video clips with associated QA-pairs, spanning four daily scenarios and 31 task types. This design makes EgoProceVQA both diverse enough to cover realistic procedural variation and structured enough to support systematic benchmarking.

Beyond benchmark construction, we also explore how agents can effectively tackle these procedural challenges. Existing procedural learning methods often rely on vision-based clustering or sequence extraction over multiple videos of the same task \cite{bansal2022myview,mahmood2026procedure,chowdhury2024opel}, which is less suitable for flexible assistants expected to reason efficiently under limited observations and varying task configurations. Other methods depend on supervised training or reinforcement-learning-based optimization for tool use \cite{shah2023steps,lin2022learning,liu2025longvideoagent,vinod2025egovlm}, which can be resource-intensive and task-specific. Therefore, we propose \textbf{EgoProceAgent}, a training-free, self-skill-exploration agent inspired by OpenClaw. EgoProceAgent uses a general-purpose procedural-understanding tool library and explicit subskill specifications, and performs four-stage self-exploration to identify effective skill strategies for different question types without ground-truth feedback. Overall, our contributions in this paper are summarized as follows:
\begin{itemize}
    \item We propose \textbf{EgoProceVQA}, a new task for systematically evaluating egocentric procedural understanding in MLLMs and agents through six progressively structured, key-step-centric subtasks as shown in Fig.~\ref{fig1}.
    \item We develop \textbf{EgoProceGen}, an automatic data generation platform that enables scalable, controllable, and transferable construction of our procedural QA benchmark across diverse activity scenarios.
    \item  We establish a large-scale benchmark with 3,600 egocentric video clips and associated QA-pairs across four daily scenarios and 31 task types, together with an evaluation protocol for fine-grained diagnosis of procedural reasoning abilities.
    \item We provide comprehensive benchmark results, including a strong training-free agentic baseline (\textbf{EgoProceAgent}), and show that current systems still face substantial challenges in key-step-level procedural understanding.

\end{itemize}

\section{Related Work}
% ----------------------------------------------------------
\subsection{Egocentric Datasets and Benchmarks}
% ----------------------------------------------------------

With the advent of Ego4D \cite{grauman2022ego4d}, egocentric video understanding has gained increasing attention. Early benchmarks like EgoVQA \cite{fan2019egovqa}, EgoTaskQA \cite{jia2022egotaskqa}, and EgoThink \cite{cheng2024egothink} primarily evaluate simple visual perception tasks such as object and action recognition, which limits their ability to evaluate deep reasoning \cite{su2025annexe,wang2025eva02,su2025care,su2026earl} or high-level task-oriented performance shown in Table~\ref{tab1}. To address this limitation, recent works like EgoPlan-Bench \cite{chen2023egoplan} and ProMQA \cite{hasegawa2025promqa} have shifted toward complex planning and procedural learning. However, existing datasets, like EgoProceL \cite{bansal2022myview}, often rely exclusively on visual features for key-step sequence recognition, making them sub-optimal for directly evaluating modern Multi-modal Large Language Models (MLLMs). To address this gap, we propose a novel benchmark featuring complex procedural QA tasks tailored for MLLMs and agents. Unlike prior visual-centric datasets, our work explicitly assesses key-step-level reasoning (e.g., step ordering and missing step detection), which requires joint understanding of both textual instructions and visual inputs.

% ----------------------------------------------------------
\subsection{Key Step Level Procedural Understanding}
% ----------------------------------------------------------

Key-step-level understanding is crucial for comprehending procedural tasks and providing procedural assistance \cite{li2025building}. Some prior works employ visual feature clustering methods \cite{chowdhury2024opel} to perform key-step sequence recognition by leveraging multiple videos of the same task. However, in egocentric AI assistant scenarios such as AI glasses, models often encounter cross-modal single-video processing tasks that extend beyond mere key-step sequence recognition. Consequently, datasets designed for pure visual methods, such as EgoProceL, struggle to support comprehensive key-step-level evaluation and lack effective annotations for the QA settings commonly used in MLLM evaluation. Our work not only conducts a thorough assessment through six key-step-level tasks spanning four scenarios and 31 task categories, but also evaluates the model's cross-modal information processing and understanding capabilities.

\subsection{Agent Skills}

With the rising prominence of OpenCLaw, the concept of agent skills \cite{xu2026agent,hao2025uncertainty,li2026guic2coarsetofineguigrounding} has attracted increasing attention, leading to emerging work on their evaluation and generation \cite{li2026skillsbench}. In this context, we propose a method that, given a task, autonomously explores and learns an optimal skill policy to solve it, offering strong flexibility and adaptability.

\begin{figure*}
    \centering
    \vspace{-5pt}
    \includegraphics[width=0.8\linewidth]{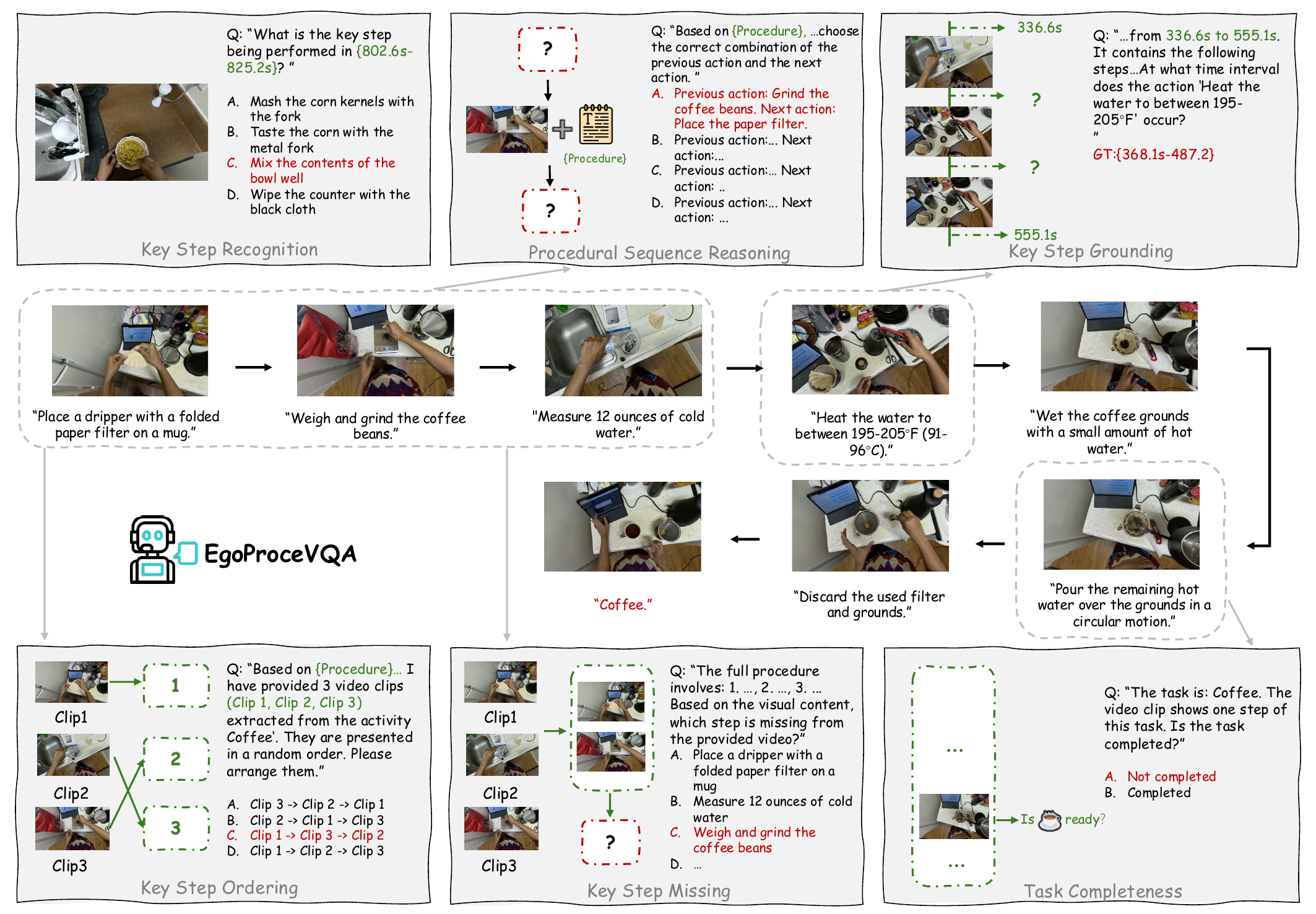}
\vspace{-7pt}
    \caption{Overview of EgoProceVQA. For each task type, we present a visualized example to facilitate understanding. At the center of the figure, we show a complete key-step sequence for preparing coffee, around which all examples except KSR are constructed. Our task design remains tightly focused on the key-step sequence of the procedural task.}
    \label{fig2}
    \vspace{-7pt}
    
\end{figure*}

\section{Benchmark, Platform, and Metrics}

In this section, we first describe the construction of EgoProceVQA, including data collection, benchmark design, and automatic QA generation. We then introduce the evaluation metrics used to assess model performance across different procedural reasoning tasks.

\subsection{Data Collection}

To build EgoProceVQA, we first collect egocentric procedural videos from diverse real-world activity domains. We employ four egocentric procedural task datasets as our video sources: \textbf{CaptainCook4D} \cite{peddi2024captaincook4d}, which covers 24 kitchen recipes; \textbf{EPIC-Tent} \cite{damen2020epic}, which focuses on outdoor tent setup; \textbf{Assembly101} \cite{sener2022assembly101}, which involves indoor toy car assembly; and \textbf{EgoOops} \cite{haneji2025egooops}, which comprises five types of tabletop crafting tasks such as cardboard handiwork. Considering potential training set construction in future research, we perform proportional random sampling based on the original scale of each dataset to construct four scenarios in EgoProceVQA. Specifically, we select 632 video clips for the \textbf{Cooking} category (yielding 1,800 QA pairs), 124 clips for \textbf{Outdoor Tent Assembly} (600 QA pairs), 341 clips for \textbf{Toy Car Assembly} (600 QA pairs), and 175 clips for \textbf{Handicraft Activities} (600 QA pairs).

\subsection{Benchmark Design}

Based on the collected videos and step annotations, we formulate EgoProceVQA as a key-step-centric benchmark for egocentric procedural understanding. Specifically, EgoProceVQA comprises six task types, each targeting a distinct level of procedural competence. Fig.~\ref{fig2} presents one representative example for each task type.

\paragraph{Task 1 — Key Step Recognition (KSR)}
Input: a single clip $v_1$.
Question sample: "What is the key action performed in this clip?"
Answer: one of four options (A-D); the correct choice is the
ground-truth key step description.
This task provides the perceptual grounding foundation of key-step-level reasoning, upon which all
higher-order reasoning rests.

\paragraph{Task 2 — Procedural Sequence Reasoning (PSR)}
Input: a clip $v_1$ plus the full procedure context
$\mathcal{P}$.
Question sample: "Given the shown clip as the current step, what are the
immediately preceding and following steps?''
Answer: one of four combined \textsc{(prev, next)} option pairs.
This task examines the model’s understanding of commonsense causal chains and short-horizon temporal prediction, as well as its ability to align procedural text with visual information.

\paragraph{Task 3 — Key Step Grounding (KSG)}
Input: a composite clip $v_1$ spanning 3-5 consecutive key steps,
accompanied by a numbered step list.
Question sample: "At what time interval does key step \textit{Z} occur?''
Answer: an open time interval $(t_s,t_e)$ in seconds.
This task evaluates the model’s ability to identify key steps and perform temporal grounding in an open-ended setting, which is crucial because robust key-step-level understanding requires not only recognizing the key steps within a segment but also capturing their temporal information to enable more effective procedural assistance.

\paragraph{Task 4 — Key Step Ordering (KSO)}
Input: three clips $\{v_1,v_2,v_3\}$ in randomised order, plus procedure context $\mathcal{P}$.
Question sample: ''Arrange the clips into the correct chronological order according to the procedure context.''
Answer: one of four permutation options.
We require the model to possess the ability to infer the ordering of relevant key steps within a segment based on the procedure text. Successfully addressing this challenging problem can equip the model with the capacity to tackle more advanced procedural error detection and verification tasks.

\paragraph{Task 5 — Key Step Missing (KSM)}
Input: two clips $(v_1,v_2)$ representing steps 1 and 3 of a
three-step procedural sequence.
Question sample: "Which step is missing from the provided video?''
Answer: one of four options; the correct answer is step~2's
description. (Correct answer is random.)
The model is not told which positions the clips occupy; it must infer the gap
from visual content and procedural context alone.

\paragraph{Task 6 — Task Completeness (TC)}
Input: a single clip $v_1$ and the task name.
Question: ``Is the task completed after this step?''
Answer: binary choice (Completed / Not completed).
This task mirrors the completion judgment required by proactive
assistants~\cite{proassist2025}.

% ──────────────────────────────────────────────────────────────────
\subsection{EgoProceGen}
% ──────────────────────────────────────────────────────────────────
To support scalable and controllable construction, we develop EgoProceGen, an automatic generation platform for procedural QA data. EgoProceGen integrates automated QA construction with a dual-pipeline generation mechanism to produce task instructions and associated QA pairs across all six evaluation dimensions. Specifically, it uses an LLM-assisted semantic pipeline for tasks requiring semantically confounded distractors (Tasks 1 and 2), and a rule-based structural pipeline for tasks whose ground truth can be deterministically derived from temporal metadata (Tasks 3–6).

\subsubsection{LLM-Assisted Semantic Generation (Tasks 1, 2)}

For tasks with semantically confounded distractors, EgoProceGen employs an
LLM-in-the-loop mechanism.
Video frames are sampled at 1\,frame/s, resized to 512\,px width, and encoded
as base64 images.
Together with the ground-truth step description, these inputs are passed to
Qwen3.5-Plus via a structured prompt that instructs the model to generate distractors according to the following confusion strategies: 

\textbf{Action confusion:} same object, different action (e.g., \textit{taking} $\to$ \textit{washing}); \textbf{Object confusion:} same action, different object (e.g., \textit{stirring spoon} $\to$ \textit{cutting knife}); \textbf{Temporal confusion:} plausible steps drawn from adjacent positions in the canonical procedure.

To prevent hallucination, metadata fields (\texttt{video\_id},
\texttt{start\_time}, \texttt{end\_time}) are forcibly rewritten after parsing
the model's JSON output.
All generated items are subsequently reviewed by three human annotators, who
discard or revise any option that fails to probe procedural understanding or
introduces an ambiguous ground truth.

\subsubsection{Rule-Based Structural Generation (Tasks 3-6)}

Tasks whose ground truth is structurally determined are handled by a fully
deterministic pipeline that requires no LLM involvement, ensuring complete
reproducibility and eliminating model-induced variance.
Each task type is handled by a dedicated generation module, details can be found in Appendix.

% ──────────────────────────────────────────────────────────────────
\subsection{Evaluation Metrics}
% ──────────────────────────────────────────────────────────────────
Traditional open-ended evaluation for procedural video understanding inherits
two established failure modes: models can produce plausible yet factually
incorrect answers that are difficult to verify automatically, and
LLM-as-a-Judge pipelines introduce additional variance and
cost \cite{cheng2024egothink}.
To overcome these limitations, EgoProceVQA adopts a
closed-set multiple-choice format for five of the six tasks, enabling
deterministic, reproducible evaluation without human or LLM judges.
Each question targets a precisely defined procedural event with
semantically proximate distractors, preventing models
from exploiting surface-level cues and forcing genuine procedural reasoning. We choose Accuracy for multiple-choice task 1, 2, 4, 5, 6, and tIoU \cite{cai2024temporalbench} for temporal grounding task 3 in order to ensure a fair and objective evaluation. Definitions can be found in Appendix.

\begin{figure*}
\vspace{-7pt}
    \centering
    \includegraphics[width=0.9\linewidth]{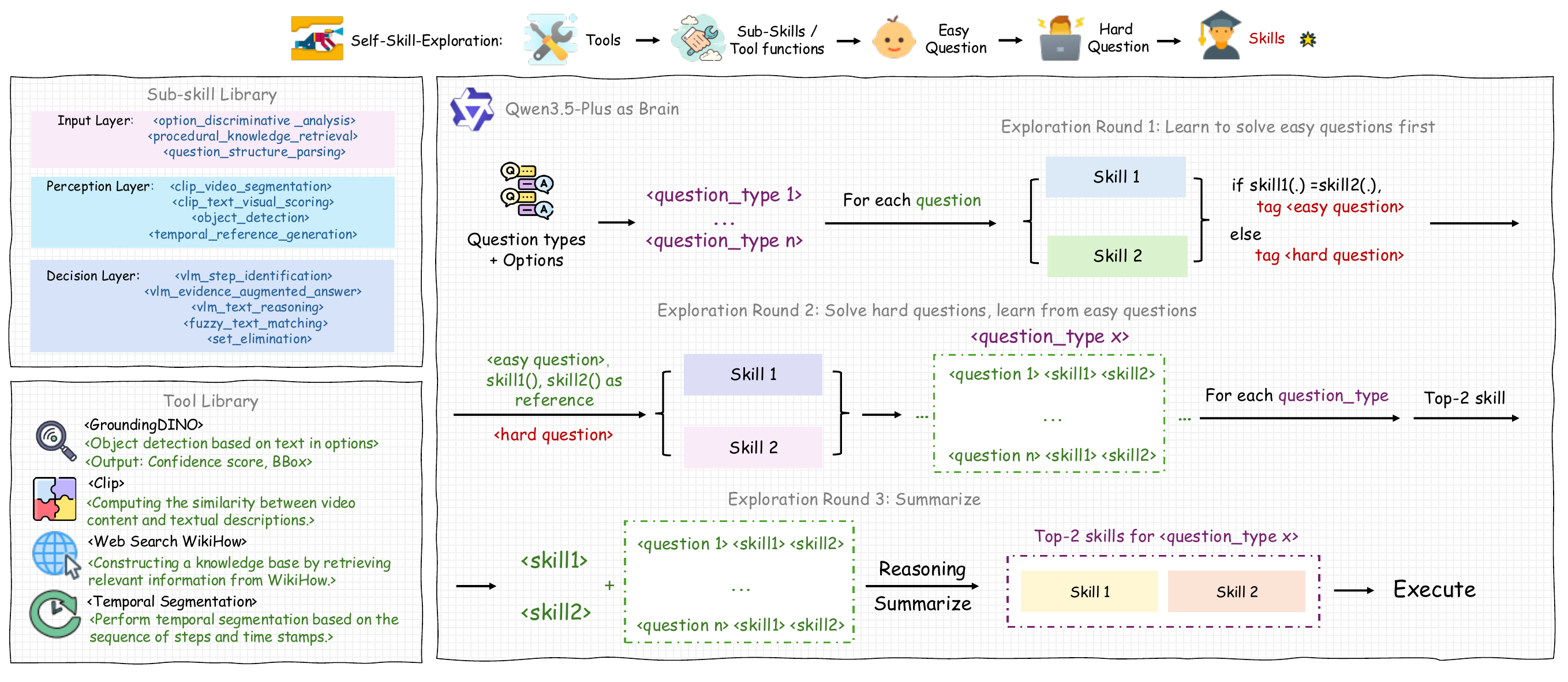}
\vspace{-8pt}
    \caption{Overview workflow of EgoProceAgent, Skill1(.) represents answers of skill 1. Based on the tool usage defined for each sub-skill, our self-skill-exploration proceeds in three main stages: it first learns to solve "easy questions" with generated skill strategies, then leverages these solutions as references to tackle "hard questions", and finally distills the optimal strategy for this class of problems.}
    \label{fig3}
    \vspace{-8pt}
\end{figure*}

\section{Methodology: EgoProceAgent}

Procedural understanding often requires different reasoning strategies for different question types, much like humans selectively apply different problem-solving skills depending on the task. For example, when understanding a video with multiple key steps, humans often first decompose it into key procedural units. When locating a missing key step, they may rely on elimination rather than direct selection. Therefore, the key challenge for an agent is not only tool invocation, but also the selection and composition of appropriate sub-skills (tool functions). To address this challenge, we propose EgoProceAgent, a training-free agentic framework that performs self-skill-exploration over a procedural-understanding tool library. EgoProceAgent progressively explores how to combine basic functions for solving simpler tasks, and then distills reusable skill strategies that can generalize to more complex procedural reasoning problem.

\subsection{Framework Overview}

EgoProceAgent consists of two clearly separated phases, self-skill-exploration phase is shown in Fig.~\ref{fig3}.

\textbf{Phase~I: Self-Skill-exploration (Offline).} A powerful LLM functions as the central “brain”: it first categorizes all questions and their corresponding option types into several sub-types, then performs self-skill-exploration of its optimal skill strategy over all questions, and ultimately distills a generalizable skill corresponding to each sub-type. \textbf{Phase~II: Skill Execution (Online).} Each incoming question is classified into its sub-type, and the corresponding distilled skill strategy is executed by the video-LLM $\mathcal{M}_{\text{exec}}$, sequentially invoking the required sub-skills and tools. If the primary strategy yields no answer, the fallback strategy is automatically attempted.

This separation yields two key advantages.
First, it decouples strategic reasoning (which sub-skills to invoke and in what order) from visual reasoning (interpreting video frames to execute each sub-skill), enabling the use of a powerful but vision-agnostic LLM as planner and a smaller but video-capable video-LLM as executor, each model operates where its strengths lie. Second, the self-skill-exploration is highly flexible and automated, and can it rapidly adapt to new procedural understanding tasks.

\subsection{Tool Library}

EgoProceAgent maintains a shared Tool Library consisting of four specialized tools. Each tool is lazily initialized upon first invocation to minimize memory overhead.

\textbf{Grounding DINO}~\cite{liu2024grounding} enables open-vocabulary object detection conditioned on text queries. \textbf{CLIP}~\cite{radford2021learning} provides a shared embedding space for images and text, enabling zero-shot visual-semantic matching. We employ CLIP in two modes: \textbf{Option Scoring:} Compute the similarity between the option texts and the visual content. And \textbf{Boundary Detection:} Segment the video by exploiting visual discontinuities between different clips. \textbf{Web search WikiHow} provides Web Knowledge Retrieval that queries WikiHow\footnote{\url{https://www.wikihow.com}} for structured step-by-step procedure descriptions. \textbf{Temporal Segmentation} generates structured time reference anchors based on the question. Details of the tools can be found in Appendix.

\subsection{Sub-Skill Decomposition}

Most agentic frameworks now treat tool calling as a fixed module, and seek to enhance performance on specific tasks by providing additional information. While effective, the fixed coupling between tools and strategies limits adaptability, the same tool may serve different roles across question types. To enable composable, question-adaptive reasoning, we decompose the tool calling into \textbf{12 atomic sub-skills (tool functions)} organized in a three-layer architecture:

 \textbf{Input Layer} ($\mathcal{I}$): extracts structured signals from the question and options.
    \begin{itemize}[leftmargin=1em, itemsep=1pt]
        \item $I_1$: Option Discriminative Analysis: parses option differences (action vs.\ object vs.\ quantity) and generates a focus hint.
        \item $I_2$: Procedural Knowledge Retrieval: loads procedure text (from question), or WikiHow (from Web search) entries based on type-specific access rules.
        \item $I_3$: Question Structure Parsing: extracts target step, listed steps, and clip temporal boundaries.
    \end{itemize}
 \textbf{Perception Layer} ($\mathcal{P}$): processes video frames using external tools.
    \begin{itemize}[leftmargin=1em, itemsep=1pt]
        \item $P_1$: CLIP Video Segmentation: detects clip boundaries via inter-frame similarity valleys.
        \item $P_2$: CLIP Text-Visual Scoring: computes option, frame cosine similarity rankings.
        \item $P_3$: GroundingDINO Object Detection: open-vocabulary object detection on key-frames.
        \item $P_4$: Temporal Reference Generation: produces uniform temporal anchors for temporal grounding.
    \end{itemize}
 \textbf{Decision Layer} ($\mathcal{D}$): produces the final answer.
    \begin{itemize}[leftmargin=1em, itemsep=1pt]
        \item $D_1$: VLM Step Identification: identifies which procedure step a video segment depicts.
        \item $D_2$: Evidence-Augmented VLM Answer: injects all accumulated evidence into a VLM call that directly produces the answer.
        \item $D_3$: Text Reasoning: VLM-based text-only reasoning over identified steps and procedure order.
        \item $D_4$: Fuzzy Text Matching: maps identified steps to option letters via sequence similarity.
        \item $D_5$: Set Elimination: determines the missing step by set difference.
    \end{itemize}

A skill strategy $\pi = (s_1, s_2, \ldots, s_L)$ is an ordered sequence of sub-skill IDs drawn from $\mathcal{S} = \mathcal{I} \cup \mathcal{P} \cup \mathcal{D}$, where each sub-skill's output feeds into subsequent sub-skills via a shared context dictionary. A valid strategy must satisfy: (i) $|\pi| \geq 2$, (ii) at least one sub-skill from each layer, and (iii) the terminal sub-skill $s_L \in \{D_2, D_3, D_4\}$ must produce a final answer.

\subsection{Self-Skill-Exploration}

A key challenge is determining the optimal strategy $\pi^*$ for each question without access to ground-truth labels. We address this with a four-pass self-exploration protocol that discovers effective strategies through consistency-based self-learning.

Given a question dataset  $\mathcal{Q} = \{(q_i, \mathcal{O}_i, \mathcal{V}_i)\}_{i=1}^{N}$, we assume access to a planning model $\mathcal{M}_{\text{plan}}$ (a strong text LLM) and an execution model $\mathcal{M}_{\text{exec}}$ (a local video-LLM). The four passes are summarized in Algorithm 1 shown in Appendix.

\subsubsection{Pass~0: Sub-Type Classification}

$\mathcal{Q}$ is divided into batches of size $B$ (we use $B{=}200$), and the planning model is prompted to assign each question a short categorical label based solely on its textual structure and answer format:
\begin{equation}
    \sigma(q_i) = \mathcal{M}_{\text{plan}}\!\left(\texttt{classify}\bigl(q_i, \mathcal{O}_i\bigr)\right), \quad \forall\, q_i \in \mathcal{Q}.
\end{equation}
The planning model is instructed to produce a concise type name (eg., \texttt{action\_recognition}, \texttt{step\_ordering}, \texttt{temporal\_grounding}) that captures the reasoning pattern required. After all batches are classified, we apply a label normalization step, merging near-duplicate labels via edit-distance matching, to obtain a consolidated label set $\Lambda = \{\lambda_1, \lambda_2, \ldots, \lambda_L\}$. Each question is thereby assigned to exactly one discovered type $\sigma(q_i) \in \Lambda$, partitioning $\mathcal{Q}$ into $L$ groups that share similar structural patterns and are expected to benefit from the same reasoning strategy.

\subsubsection{Pass~1: Dual-Strategy Exploration}

For each discovered type $\lambda$, we apply dual-strategy skill-exploration to all questions in that group, denoted $\mathcal{Q}_\lambda = \{q_i : \sigma(q_i) = \lambda\}$. For each question $q \in \mathcal{Q}_\lambda$, the planning model proposes two candidate strategies conditioned on type-specific constraints $\mathcal{C}_\lambda$ (automatically inferred from the discovered type's characteristics):
\begin{equation}
    (\pi^A, \pi^B) = \mathcal{M}_{\text{plan}}\!\left(\texttt{plan}\bigl(q, \mathcal{O};\, \mathcal{C}_\lambda\bigr)\right).
\end{equation}
Both strategies are executed by $\mathcal{M}_{\text{exec}}$ to obtain answers:
\begin{equation}
    a^A = \texttt{Execute}(\pi^A, q, \mathcal{V}), \quad a^B = \texttt{Execute}(\pi^B, q, \mathcal{V}).
\end{equation}
A sample is marked as <easy question> if the two answers agree ($a^A = a^B \neq \varnothing$), and its strategy pair is recorded in a strategy memory $\mathcal{M}_\lambda$, otherwise as <hard question> since it leads to a discrepancy under the two strategies.

The sub-type-specific constraints $\mathcal{C}_\lambda$ encode structural knowledge about each discovered question type: eg., types involving procedure text (question provides) should include $I_2$ (knowledge retrieval); types requiring temporal localization should include $P_5$ (temporal reference); the terminal skill must produce a final answer. These constraints are derived from the type description discovered in Pass~0 and ensure that the planner's exploration stays within a valid and effective search space. A repair function automatically injects missing mandatory skills.

\subsubsection{Pass~2: Reference-Guided Consolidation}

The <hard question> samples from Pass~1 are re-planned using <easy question> strategies from the same discovered type as reference. For an <hard question> $q$ with type $\lambda$:
\begin{equation}
    (\pi^A, \pi^B) = \mathcal{M}_{\text{plan}}\!\left(\texttt{plan}\bigl(q, \mathcal{O};\, \mathcal{C}_\lambda,\, \mathcal{M}_\lambda\bigr)\right),
\end{equation}
where $\mathcal{M}_\lambda$ provides up to 5 reference strategies from <easy question> marked questions. Strategy $\pi^A$ and $\pi^B$ are executed, and the results are added to the strategy memory.

\subsubsection{Pass~3: Strategy Distillation}

After Passes 1 and 2, each discovered type $\lambda$ has accumulated a set of strategy records in $\mathcal{M}_\lambda$. We distill a primary and secondary strategy pair $(\pi_\lambda^*, \pi_\lambda^{**})$ via frequency-based voting followed by LLM verification:
\begin{equation}
    \pi_\lambda^* = \arg\max_{\pi} \sum_{r \in \mathcal{M}_\lambda} \mathbb{1}[\pi \in r], \quad
    \pi_\lambda^{**} = \arg\max_{\pi \neq \pi_\lambda^*} \sum_{r \in \mathcal{M}_\lambda} \mathbb{1}[\pi \in r],
    \label{eq:distill}
\end{equation}
where $r$ ranges over all records and each record contributes its two strategies. The top candidates are then presented to $\mathcal{M}_{\text{plan}}$ alongside representative questions for confirmation or minor refinement, yielding the final strategy card per discovered type.

\subsubsection{Deployment}

For full evaluation, each question $q_i$ is mapped to its discovered type $\sigma(q_i)$, and the corresponding primary strategy $\pi_{\sigma(q_i)}^*$ is executed. If the primary strategy yields an empty answer, the secondary strategy $\pi_{\sigma(q_i)}^{**}$ serves as fallback:
\begin{equation}
    a_i = \begin{cases}
        \texttt{Execute}\bigl(\pi_{\sigma(q_i)}^*, q_i, \mathcal{V}_i\bigr), & \text{if result} \neq \varnothing, \\
        \texttt{Execute}\bigl(\pi_{\sigma(q_i)}^{**}, q_i, \mathcal{V}_i\bigr), & \text{otherwise}.
    \end{cases}
\end{equation}

\begin{table*}[t]
    \centering
    \vspace{-5pt}
      \caption{Random means results of randomly selected. Avg represents the average of all accuracy metrics except KSG. >=20s means samples' acc (exp KSG) longer than or equal to 20s, >=20s (G) means samples' tIoU longer than or equal to 20s. Red means best performance 7B/8B model in 8-frame input, purple means best performance 7B/8B model in 16-frame input.}
    \label{tab2}
      \vspace{-5pt}
    \scalebox{0.76}{
    \begin{tabular}{ccccccccccccc}
    \toprule
         Model & Frame & KSR & PSR & KSG & KSO & KSM & TC   & >=20s(G)&<20s(G)& >=20s&<20s& Avg  \\
    \midrule
       Random & - & 25.0 & 25.0 & - & 25.0 & 25.0 & 50.0   & -&-& -&-& 30.0 \\
\midrule
  Human & - & 91.0 & 91.0 & 0.85 & 93.0 & 96.0 & 90.0   & -&-& -&-& 92.2 \\
\midrule
 \multicolumn{13}{c}{Closed-source}\\
\midrule
  GPT-4o & 8 & 59.3 & 44.3 & 0.27 & 25.7 & 52.3 & 51.5   &0.24 &0.36& 46.3&47.2& 46.6 \\
  \midrule
  GPT-5.1  & 8 & 66.5 & 47.5 & 0.34 & 26.5 & 55.2 & 62.5&0.32&0.39&53.3&49.0&51.6\\
  \midrule
       Gemini-3-Flash  & 8 & 62.7 & 41.2 & 0.34 & 24.3 & 69.7 & 58.8&0.32&0.42&49.8&53.8&51.3\\
  \midrule
  \multirow{2}{*}{Qwen3.5-Plus (Commercial)}  & 8 & 76.0 & 86.3 & 0.32 & 38.2 & 59.5 & 58.5   & 0.40&0.27&67.4 &57.8& 63.7  \\
                                                & 16 & 76.5 & 89.2 & 0.28 & 37.5 & 63.0 & 59.0   &0.37 &0.23&70.6 &56.2& 65.0  \\
\midrule
 \multicolumn{13}{c}{Open-source}\\
\midrule
        \multirow{2}{*}{LLaVA-OneVision} & 8  & 52.5 & 39.2 & 0.10 & 26.8 & 11.2 & 52.3   & 0.07&0.18& 36.3&36.6& 36.4 \\
                                         & 16 & 52.0 & 39.0 & 0.10 & 26.2 & 9.2 & 52.7   &0.07 &0.18&35.8 &35.8& 35.8 \\
    \midrule
     \multirow{2}{*}{Vinci} & 8  & 41.2 & 40.0 & 0.20 & 27.2 & 23.0 & 52.5   &0.18 &0.26&35.8 &38.3& 36.8 \\
                                         & 16 & 41.7 & 38.5 & 0.20 &25.0  & 23.0 & 52.5   &0.18 &0.26& 35.5&37.2& 36.1 \\
        \midrule
         \multirow{2}{*}{EgoGPT} & 8  & 49.5 & 38.3 & 0.10 & 26.7 & 4.3 & \cellcolor{red!10}53.7   & 0.07&0.17& 33.9&35.4& 34.5 \\
                                         & 16 & 49.5 & 38.3 & 0.08 & 26.5  & 5.2 & 52.2   &0.05 &0.15&33.9 &35.0& 34.3 \\
        \midrule
         \multirow{2}{*}{Video-LLaMA2} & 8  & 34.2 & 40.5 & 0.17 & 26.8 & 42.8 & 52.5   &0.14 &0.24&38.6 &40.6& 39.4 \\
                                         & 16 & 34.3 & 40.7 & 0.16 &25.7  & 41.7 & 52.5   &0.13 &0.25& 38.9&39.1& 39.0 \\
        \midrule
       Video-LLaVA  &8 & 29.5 & 27.7 & 0.15 &21.3& 23.8 & 49.8   & 0.12&0.24& 27.8&34.6& 30.4 \\
        \midrule
        \multirow{2}{*}{Qwen2-VL-7B}   & 8  &  55.2 & 43.8 & 0.22 & 24.8 & 38.7 & 52.5   &0.20 &0.26&41.1 &46.0& 43.0  \\
                                                & 16 & \cellcolor{blue!10} 57.0 &  43.0 &  0.22 & 24.3 & 42.0 & 53.0   &0.20 &0.27&41.7 &47.2&  43.9  \\
        \midrule
        \multirow{2}{*}{Qwen2.5-VL-7B} & 8  & 44.3 & 39.5 & 0.20 & 22.2 & 42.2 & 53.5   &0.18 &0.25&39.1 &42.2& 40.3  \\
                                                & 16 & 51.2 & 40.2 & 0.18 & 24.0 &  45.8 &  \cellcolor{blue!10} 53.3   &0.15 &0.24& 42.6&43.5& 42.9  \\
         \midrule
         \multirow{2}{*}{Qwen3-VL-8B} & 8  & \cellcolor{red!10}59.0 & 45.7 & 0.24 & 27.2 & 25.3 & 52.5   & 0.20&0.34& 42.1&41.7& 41.9 \\
                 & 16 & 55.0 & 47.8 & 0.24 &\cellcolor{blue!10}27.8  & 24.7 & 52.8   &0.20 &0.34&41.4 &42.1& 41.6 \\
         \midrule
  \multirow{2}{*}{InternVL3-38B}  & 8  & 54.2 & 47.6 & 0.15 & 28.7 & 51.5 & 61.4   &0.37&0.25&66.1&57.0&  48.7 \\
                                                & 16 & 62.8 & 49.3 & 0.13 & 26.3 & 56.3 & 61.8 &0.39 &0.25&69.5&56.4& 51.3 \\ 
        \midrule
        \multirow{2}{*}{Ours (Qwen2.5-VL-7B)}& 8 & 49.2 $\textcolor{green}{\uparrow{\scriptsize 11.1\%}}$ & \cellcolor{red!10}72.8 $\textcolor{green}{\uparrow{\scriptsize 84.3\%}}$ & \cellcolor{red!10}0.29 $\textcolor{green}{\uparrow{\scriptsize 45.0\%}}$ & \cellcolor{red!10}27.8 $\textcolor{green}{\uparrow{\scriptsize 25.2\%}}$ & \cellcolor{red!10}43.2 $\textcolor{green}{\uparrow{\scriptsize 2.4\%}}$ & 53.2   &0.27 &0.37& 52.0&44.9&  \cellcolor{red!10}49.2 $\textcolor{green}{\uparrow{\scriptsize 22.1\%}}$ \\ 
        & 16 & 54.5 $\textcolor{green}{\uparrow{\scriptsize 6.4\%}}$ & \cellcolor{blue!10}81.2 $\textcolor{green}{\uparrow{\scriptsize 102.0\%}}$ & \cellcolor{blue!10}0.29 $\textcolor{green}{\uparrow{\scriptsize 61.1\%}}$ & \cellcolor{blue!10}27.8 $\textcolor{green}{\uparrow{\scriptsize 15.8\%}}$ & \cellcolor{blue!10}48.2 $\textcolor{green}{\uparrow{\scriptsize 5.2\%}}$  & \cellcolor{blue!10}53.3   & 0.27&0.36& 55.5&49.0&  \cellcolor{blue!10}53.0 $\textcolor{green}{\uparrow{\scriptsize 23.5\%}}$ \\ 
    \bottomrule
    \end{tabular}
   }
  
    \vspace{-3pt}
\end{table*}

\section{Experiment}

\subsection{Experiment Setting}

\textbf{Model.} We select two proprietary and eight open-source models for a comprehensive and representative evaluation. For proprietary models, we adopt GPT-4o \cite{hurst2024gpt}, widely recognized as a recent state-of-the-art MLLM in many benchmarks, and the closed-source version of Qwen3.5-Plus as strong references. For open-source general video-LLMs, we include LLaVA-OneVision \cite{li2024llava}, Video-LLaVA \cite{lin2024video}, and Video-LLaMA2 \cite{cheng2024videollama}. EgoGPT \cite{yang2025egolife} and Vinci \cite{huang2025vinci}, which are fine-tuned on egocentric data, are chosen to represent egocentric video agents. As widely used baseline models, we also evaluate Qwen2-VL-7B (-Instruct) \cite{wang2024qwen2}, Qwen2.5-VL-7B (-Instruct) \cite{qwenteam2025qwen25vl} and Qwen3-VL-8B (-Instruct) \cite{qwenteam2025qwen3vl}. To showcase the performance of large open-source models, we further include InternVL3-38B \cite{zhu2025internvl3}.

\textbf{Experiment.} We conducted uniform sampling with 8 and 16 frames as inputs on a single RTX 4090 GPU (7B/8B). The decoding was configured with `do\_sample = False` to ensure deterministic and controllable outputs. EgoProceAgent's settings are kept consistent with the baseline.

\subsection{Benchmark Results}

Our experimental setup enables fair comparisons under matched input frame conditions (8 vs 8, 16 vs 16), and also allows us to examine how increasing the number of input frames affects procedural understanding. The results show that most open-source 7B models exhibit weak key-step-level understanding, substantially lagging behind human performance. With 8 input frames, the best overall performance is achieved by the closed-source Qwen3.5-Plus (63.7). Among open-source large models, InternVL3-38B reaches 48.7, surpassing GPT-4o (46.6) and ranking second. With 16 input frames, Qwen3.5-Plus again performs best (65.0). For the more challenging KSO task with 8 frames, Qwen3-VL-8B achieve the highest score of 27.2, which is still close to random performance. Even the strong closed-source GPT-4o only attains 25.7. In KSM, model performance varies substantially, among 8-frame open-source 7B models, Qwen2.5-VL reaches 42.2, while the 38B InternVL3 achieves 51.5. For relatively simpler tasks such as KSR and PSR, most models perform comparatively well. InternVL3 (16-frame) attains the best open-source results of 62.8 and 49.3, respectively. In contrast, for TC, which probes whether a model can determine task completion based on pretrained knowledge, most models perform only slightly above random, with InternVL3 (16-frame) obtaining the best score of 61.8 shown in Table~\ref{tab2}.

Notably, the latest closed-source Qwen3.5-Plus demonstrates strong procedural understanding. It achieves the best performance on all tasks except TC, with PSR reaching as high as 89.2 (16-frame), and KSR also attaining a high score of 76.6. However, its TC performance is slightly lower than that of InternVL3, possibly because during pretraining the model was exposed to a larger number of tasks similar to those in EgoProceVQA, leading to a better understanding of task completion. The KSG results indicate that all models have poor temporal grounding ability for key steps. The best model, Qwen3.5-Plus (8-frame), only reaches 0.32. This may be due to the video frame sampling strategy, which weakens the models’ sensitivity to temporal information. Furthermore, we observe that simply increasing the number of input frames yields limited performance changes for most models and thus does not constitute a general strategy for improving procedural understanding.

We also analyze the impact of video duration on procedural understanding. In most cases, samples shorter than 20 seconds achieve clearly higher accuracy and tIoU than those with duration $\geq$ 20 seconds. However, InternVL3-38B and Qwen3.5-Plus, which exhibit the strongest procedural understanding, are exceptions: for these two models, performance on longer samples surpasses that on shorter ones. Moreover, as shown in Table~\ref{tab4}, we introduce three levels of Chain of Thought (CoT) with different lengths: level 1 corresponds to the simplest prompt "let’s think step by step", level 2 guides the model through 2–3 specified reasoning steps, and level 3 through 4–5 specified steps, in order to more comprehensively explore zero-shot reasoning enhancement. We observe that increasing the number of CoT steps and the length of the input context does not yield a consistent performance improvement or degradation on this task. Furthermore, in all experiments where CoT is introduced, the overall performance is in fact inferior to the baseline. This may be because the model already receives excessive contextual input for many tasks, and the limited capacity of the 7B model makes additional CoT reasoning an extra burden rather than a benefit.

\begin{table}
    \centering
    \vspace{-3pt}
       \caption{Performances on EgoProceL. Our performance is under 8-frame input.}
    \label{tab3}
     \vspace{-6pt}
    \scalebox{0.86}{
    \begin{tabular}{ccccc}
        \toprule
         & \multicolumn{2}{c}{PC Assembly}&\multicolumn{2}{c}{PC Disassembly} \\
         \cmidrule(lr){2-3}\cmidrule(lr){4-5}         
         & F1 & IoU & F1 & IoU\\
         \midrule
       Random  & 15.1 & 7.2 & 15.3 & 7.1\\
        Uniform & 17.4 & 8.9 & 18.1 & 9.1\\
       CnC \cite{bansal2022myview}  & 25.1 & 12.8 & 27.0 & 14.8\\
        GPL-2D \cite{bansal2024united} & 24.0 & 12.6 & 27.4 &15.9 \\
         UG-I3D \cite{bansal2024united} & 22.0 & 11.7 & 24.2 & 13.8\\
          GPL-w BG \cite{bansal2024united}& 27.6 & 14.4 &  26.9& 15.0\\
           GPL-w/o BG \cite{bansal2024united} & 27.5 & 15.2 & 26.7 & 15.2\\
           OPEL \cite{chowdhury2024opel} & \underline{33.7} & \underline{17.9} & \underline{32.2} & \underline{16.9}\\
         \rowcolor{gray!17} Ours&  \textbf{39.2}& \textbf{21.8} & \textbf{40.5} & 
\textbf{19.2}\\
           \bottomrule
    \end{tabular}
}
 
    \vspace{-16pt}
\end{table}

\subsection{EgoProceAgent Results}

Our method built upon Qwen2.5-VL-7B, achieves significant improvements over baselines across all tasks except TC, with the PSR task attaining the largest relative improvement of 84.3\% (8-frame). Among 7B/8B models, it achieves state-of-the-art performance on PSR, KSG, KSM, TC (16-frame) and average accuracy. Analysis reveals that tasks such as PSR, KSO, and KSM constitute more complex reasoning beyond KSR, requiring not only key-step recognition but also procedural judgment aligned with procedural text. Our method effectively addresses this by decoupling the reasoning process through specific skill exploration, thereby reducing reasoning burden. For the KSG task, baseline models employing uniform sampling exhibit vague temporal understanding; our method incorporates temporal reference information via skills, improving temporal reasoning capabilities. However, while leveraging tools to obtain additional reference information proves effective for KSR, it still cannot surpass Qwen3-VL and closed-source models. This is because for KSR, skill-based processing does not decouple reasoning to the same extent as for other tasks, relying more heavily on the model's intrinsic understanding. Regarding TC, we hypothesize that the model receives excessive extraneous information summarized from networks, making it challenging for a 7B model to enhance task completion understanding based solely on task names. Notably, although our method achieves comprehensive performance improvements, certain metrics still lag behind closed-source models, revealing a limitation of our approach: it fundamentally depends on the model's inherent reasoning capability, which caps the upper bound. To assess generalizability, we evaluate our method on EgoProceL shown in Table~\ref{tab3}. Given the dataset’s limitations, we use a QA framework to ensure fair comparison and perform self-skill-exploration over the defined question types. Under this skill strategy, our approach outperforms several visual clustering methods, demonstrating effective transfer to other procedural understanding tasks. The specific skill strategies derived through self-skill exploration, experiment setting of EgoProceL, and visualization results can be seen in Appendix.

\begin{table}
    \centering
    \vspace{-3pt}
     \caption{Results of ablation experiment under 8-frame input.}
    \label{tab4}
      \vspace{-6pt}
    \scalebox{0.86}{
    \begin{tabular}{ccccccc}
    \toprule
         & KSR & KSP & KSG & KSO & KSM & TC\\
         \midrule
      Baseline & 44.3 & \underline{39.5} & \underline{0.20} & 22.2 & \underline{42.2} & \textbf{53.5} \\
     CoT Level 1  & 31.8 & 26.8 & 0.06 & 20.7 & 27.2 & 50.5\\
        CoT Level 2 & 27.8 & 26.2 & 0.18 & 21.3 & 26.5 & 49.0\\
        CoT Level 3 & 42.8 & 31.2 & \underline{0.20} & 21.7 & 29.0 & 51.0 \\
        w/o Sub-Skill &  \underline{46.0}&  33.7&  \textbf{0.29}&  \underline{22.5}&  38.7& \underline{53.2}\\
        w/o Skill-Exploration & 29.3 & 29.8 & 0.05 & 16.7 & 39.3 & 39.0\\
      \rowcolor{gray!17} Ours & \textbf{49.2} & \textbf{72.8} & \textbf{0.29} & \textbf{27.8} & \textbf{43.2} & \underline{53.2}\\
        \bottomrule
    \end{tabular}
}   
    \vspace{-12pt}
\end{table}

\subsection{Ablation Study}

As shown in Table~\ref{tab4}, we conduct comprehensive ablation studies under zero-shot conditions to validate the contributions of our proposed components and compare our approach against three levels of CoT. First, we removed the sub-skill configuration and directly provided all tools as reference information when answering questions (w/o sub-skill). Next, to validate the effectiveness of our skill exploration mechanism, we replaced the learned strategies with randomly selected ones. \textbf{Impact of Sub-skill Configuration:} The results show that accuracy declines across all tasks except KSG and TC. These two tasks inherently rely on temporal information and web search content as references in our original design, thus remaining relatively stable. Overall, this decline confirms the necessity and effectiveness of the explicit sub-skill configuration. \textbf{Necessity of Skill Exploration:} For each question type, we randomly assigned a skill strategy from the pool of all available strategies (excluding the one originally explored through planning). We found that all metrics dropped below the baseline, with most falling significantly below the w/o sub-skill setting (i.e., pure tool invocation). This indicates that improper skill usage not only fails to improve performance but also actively misleads the model into incorrect reasoning, undermining the intended benefits of the tools.
\textbf{Effectiveness of Decoupled Reasoning:} Our method consistently outperforms all CoT levels across all metrics, which demonstrates that decoupling the reasoning process via a specific skill strategy is a significantly more effective method for enhancing zero-shot procedural understanding than standard CoT. More analyses can be found in Appendix.

\section{Conclusion}

We propose a novel egocentric procedural understanding task, EgoProceVQA, and establish its foundational framework from three components: a data generation platform tailored to the task’s characteristics for producing QA-pairs, a comprehensive benchmark constructed from the generated data, along with six key-step-level evaluation tasks, a self-skill-exploring agentic framework that flexibly addresses procedural understanding tasks and achieves superior performance.

\begin{acks}
The research work described in this paper was conducted in the JC STEM Lab of Machine Learning and Computer Vision funded by The Hong Kong Jockey Club Charities Trust. This research received partially support from the Global STEM Professorship Scheme from the Hong Kong Special Administrative Region.
\end{acks}

\bibliographystyle{ACM-Reference-Format}
\bibliography{references}

%\documentclass[sigconf, anonymous, review]{acmart}

%\AtBeginDocument{%
  %\providecommand\BibTeX{{Bib\TeX}}}

%%% 表格相关
%\usepackage{threeparttable}
%\usepackage{booktabs}
%\usepackage{colortbl}
%\usepackage[table]{xcolor}
%\usepackage{pifont}
%\newcommand{\cmark}{\ding{51}}
%\newcommand{\xmark}{\ding{55}}
%\definecolor{oursblue}{RGB}{220, 235, 255}
%\usepackage[table]{xcolor}
%\usepackage{colortbl}
%%% 其他工具包
%\usepackage{enumitem}
%\usepackage{multirow}
%\usepackage{algorithm}
%\usepackage{algorithmic}
%\usepackage{graphicx}
%\usepackage{url}
%\usepackage{dblfloatfix}
%\usepackage{tcolorbox}
%\tcbuselibrary{most}
%\newtcolorbox{promptbox}[1][]{colback=gray!5,colframe=gray!75,fonttitle=\bfseries,title=Prompt,#1}

%\usepackage{hyperref}
%\hypersetup{
   % colorlinks=true,
    %linkcolor=blue,
    %urlcolor=cyan,
    %bookmarks=true,
    %bookmarksopen=true,
%}

%\acmConference[MM'26]{the 34th ACM International Conference on Multimedia}{November 10-14, 2026}{Rio de Janeiro, Brazil}

%\begin{document}

\title{Supplementary Materials\\
EgoProceVQA: A Novel Egocentric Procedural Understanding Task with Self-Skill-Exploration Agent}




\maketitle



\appendix

\tableofcontents

This supplementary material provides further technical details and experimental results to complement the main manuscript. Specifically, Section~\ref{sec1} details the EgoProceGen pipeline and lists specific prompts utilized in our work, while Section~\ref{sec2} shows the specific skill strategie for each sub-type. Section~\ref{sec3} elaborates on the experimental configurations and Section~\ref{sec4} provides further information regarding the tool library. In Section~\ref{sec5}, we report experimental details on EgoProceL. Finally, Section~\ref{sec6} showcases our real-world demonstrations.

\section{Details of EgoProceGen for Task 3-6}
\label{sec1}

\textbf{Temporal Grounding.} The module selects 3-5 consecutive steps and assembles them into a
composite clip; the question text enumerates all included steps and
the answer is the ground-truth time interval of a randomly sampled target step.

\textbf{Key Step Ordering.} The module selects three consecutive steps, enumerates all six permutations,
and retains four as answer options, one correct and three incorrect, providing
a controlled combinatorial probe of step-order reasoning.

\textbf{Key Step Missing.} Given three consecutive steps $(A,B,C)$, the module withholds random step (we take step $B$ as example) and presents clips for $A$ and $C$ only; distractors are sampled from other steps
within the same procedure to maintain semantic plausibility.

\textbf{Task Completeness.} Positive samples are drawn from the \emph{final step} of each procedure;
negative samples are drawn from any non-final step, yielding a 50/50 class
balance. The order of the two options is randomised at generation time to eliminate
positional bias.

We list the prompts used by EgoProceGen in the end of the supplementary material.

\begin{algorithm}[t]
\caption{Self-Skill-Exploration for EgoProceAgent}
\label{alg:selfplan}
\begin{algorithmic}[1]
\REQUIRE Question set $\mathcal{Q}$; planning model $\mathcal{M}_{\text{plan}}$; execution model $\mathcal{M}_{\text{exec}}$
\ENSURE Strategy map $\{(\pi_\lambda^*, \pi_\lambda^{**})\}_{\lambda \in \Lambda}$

\STATE \textbf{--- Pass 0: Question-Type Discovery ---}
\FOR{each batch $\mathcal{B} \subset \mathcal{Q}$}
    \STATE $\{\sigma(q_i)\}_{q_i \in \mathcal{B}} \gets \mathcal{M}_{\text{plan}}(\texttt{classify}(\mathcal{B}))$
\ENDFOR
\STATE $\Lambda \gets \texttt{NormalizeLabels}(\{\sigma(q_i)\}_{i=1}^{N})$ \COMMENT{Edit-dist merging}
\STATE Infer constraints $\mathcal{C}_\lambda$ for each $\lambda \in \Lambda$

\STATE \textbf{--- Pass 1: Dual-Strategy Exploration ---}
\FOR{each type $\lambda \in \Lambda$}
    \STATE $\mathcal{Q}_\lambda \gets \{q_i : \sigma(q_i) = \lambda\}$
    \FOR{$q \in \mathcal{Q}_\lambda$}
        \STATE $(\pi^A, \pi^B) \gets \mathcal{M}_{\text{plan}}(\texttt{plan}(q, \mathcal{O}; \mathcal{C}_\lambda))$
        \STATE $a^A \gets \texttt{Execute}(\pi^A, q, \mathcal{V})$; \, $a^B \gets \texttt{Execute}(\pi^B, q, \mathcal{V})$
        \IF{$a^A = a^B \neq \varnothing$}
            \STATE $\mathcal{M}_\lambda.\texttt{add}(q, \pi^A, \pi^B, a^A)$ \COMMENT{<easy question>}
        \ELSE
            \STATE $\mathcal{U}_\lambda.\texttt{add}(q)$ \COMMENT{<hard question>}
        \ENDIF
    \ENDFOR
\ENDFOR

\STATE \textbf{--- Pass 2: Reference-Guided Consolidation ---}
\FOR{$q \in \mathcal{U}_\lambda$, $\forall \lambda$}
    \STATE $(\pi^A, \_) \gets \mathcal{M}_{\text{plan}}(\texttt{plan}(q, \mathcal{O}; \mathcal{C}_\lambda, \mathcal{M}_\lambda))$
    \STATE $a \gets \texttt{Execute}(\pi^A, q, \mathcal{V})$
    \STATE $\mathcal{M}_\lambda.\texttt{add}(q, \pi^A, \_, a)$
\ENDFOR

\STATE \textbf{--- Pass 3: Strategy Distillation ---}
\FOR{each type $\lambda \in \Lambda$}
    \STATE $\pi_\lambda^*, \pi_\lambda^{**} \gets \texttt{FreqVote}(\mathcal{M}_\lambda)$ \COMMENT{Eq.~\eqref{eq:distill}}
    \STATE $\pi_\lambda^*, \pi_\lambda^{**} \gets \mathcal{M}_{\text{plan}}(\texttt{verify}(\pi_\lambda^*, \pi_\lambda^{**}, \mathcal{M}_\lambda))$
\ENDFOR

\STATE \textbf{--- Deployment ---}
\FOR{$q_i \in \mathcal{Q}$}
    \STATE $a_i \gets \texttt{Execute}(\pi_{\sigma(q_i)}^*, q_i, \mathcal{V}_i)$
    \IF{$a_i = \varnothing$}
        \STATE $a_i \gets \texttt{Execute}(\pi_{\sigma(q_i)}^{**}, q_i, \mathcal{V}_i)$
    \ENDIF
\ENDFOR

\RETURN $\{(\pi_\lambda^*, \pi_\lambda^{**})\}_\lambda$, $\{a_i\}_i$
\end{algorithmic}
\end{algorithm}

\section{Specific Skill Strategies from Self-Skill Exploration}
\label{sec2}

Table~\ref{tab5} shows the specific skill strategies, and the sub-types come from the model automatically classifying all questions during the planning stage.

\begin{table}
    \centering
    \caption{Skill strategies for each sub-type.}
    \scalebox{0.8}{
    \begin{tabular}{ccc}
    \toprule
         Sub-types& Primary& Secondary\\
         \midrule
        <action\_type\_choice> & ["I1","P2","D2"] & ["I1","P3","D2"]\\
       <object\_discrimination>  & ["I1","P2","D2"],& ["I1","P3","D2"]\\
        <prev\_next\_reasoning> & ["I1","I2","D1","D4"] &["I1","I2","P3","D1","D4"] \\
       <numbered\_steps>  & ["I3","P5","D2"] &["I3","P1","P5","D2"] \\
       <temporal\_interval>  & ["I3","P5","D2"] & ["I3","P1","P5","D2"]\\
       <clip\_ordering> & ["I2","P1","D1","D3"] & ["I2","I3","P1","D1","D4","D3"]\\
      <missing\_step\_detection>  & ["I3","P1","D1","D5","D4"] & ["I3","P1","P2","D1","D5","D4"]\\
      <task\_completion> & ["I2","P2","D2"] & ["I2","P3","D2"]\\
      \bottomrule
    \end{tabular}
}
    
    \label{tab5}
    \vspace{-10pt}
    
\end{table}

\begin{figure}[t]
    \centering
    \scalebox{0.9}{
    \includegraphics[width=1\linewidth]{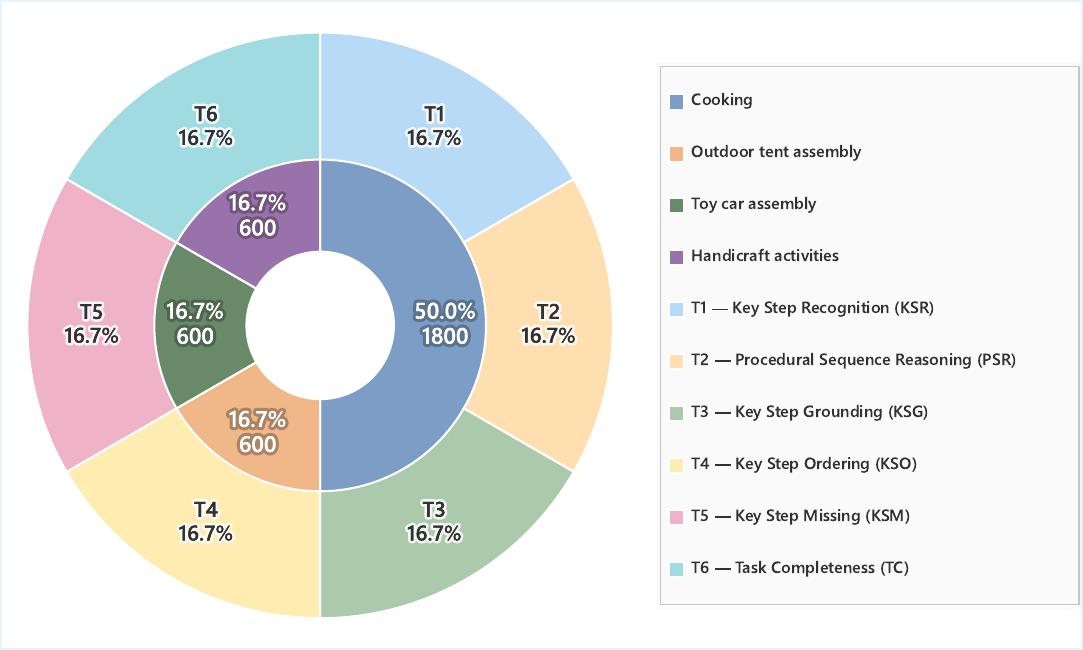}
}    \caption{Statistics of the six task types and four scenarios in EgoProceVQA.}
\vspace{-5pt}
    \label{fig4}
\end{figure}

\section{Supplementary Details for Experiment}
\label{sec3}

\subsection{Definitons of Metrics}

\paragraph{Accuracy (Tasks 1, 2, 4, 5, 6).}
For all multiple-choice tasks, the primary metric is top-1 accuracy:
\[
  \mathrm{Acc} = \frac{1}{N}\sum_{i=1}^{N}\mathbf{1}[\hat{y}_i = y_i],
\]
where $\hat{y}_i$ is the model's predicted option and $y_i$ is the
ground-truth label. An \textbf{Overall Accuracy} score (excluding Task~3) is also reported for summarising multiple-choice performance across the benchmark.

To prevent positional bias in Tasks~4 and~6, the position of the correct
answer within the option list is randomised at generation
time.

\paragraph{Temporal Intersection over Union (Task 3).}
For Temporal Grounding, we adopt the standard tIoU
metric:
\[
  \mathrm{tIoU} =
  \frac{|\hat{\mathcal{I}}\cap\mathcal{I}^*|}
       {|\hat{\mathcal{I}}\cup\mathcal{I}^*|},
\]
where $\hat{\mathcal{I}}=[\hat{t}_s,\hat{t}_e]$ is the predicted interval and
$\mathcal{I}^*=[t_s^*,t_e^*]$ is the ground-truth interval.
We report mean tIoU (mIoU), revealing whether models achieve coarse temporal
alignment or precise localisation.

\subsection{Models}

\paragraph{Closed-source Proprietary MLLMs.}
Two state-of-the-art cloud-based models have been included, which serve as reference models.

\textbf{GPT-4o} is tested using the OpenAPI.

\textbf{Qwen3.5-Plus} is accessed using Alibaba's DashScope Multimodal Conversation API, where frames have been encoded as base64-compressed JPEG images with a width of 512~px.

\paragraph{Open-source General MLLMs.}
All open-weighted models have been loaded using \texttt{bf16} precision and run using a non-thinking mode (\texttt{enable\_thinking=False}). Set do\_sample=false for all models to ensure controllable output.

\begin{figure}
    \centering
    \includegraphics[width=0.9\linewidth]{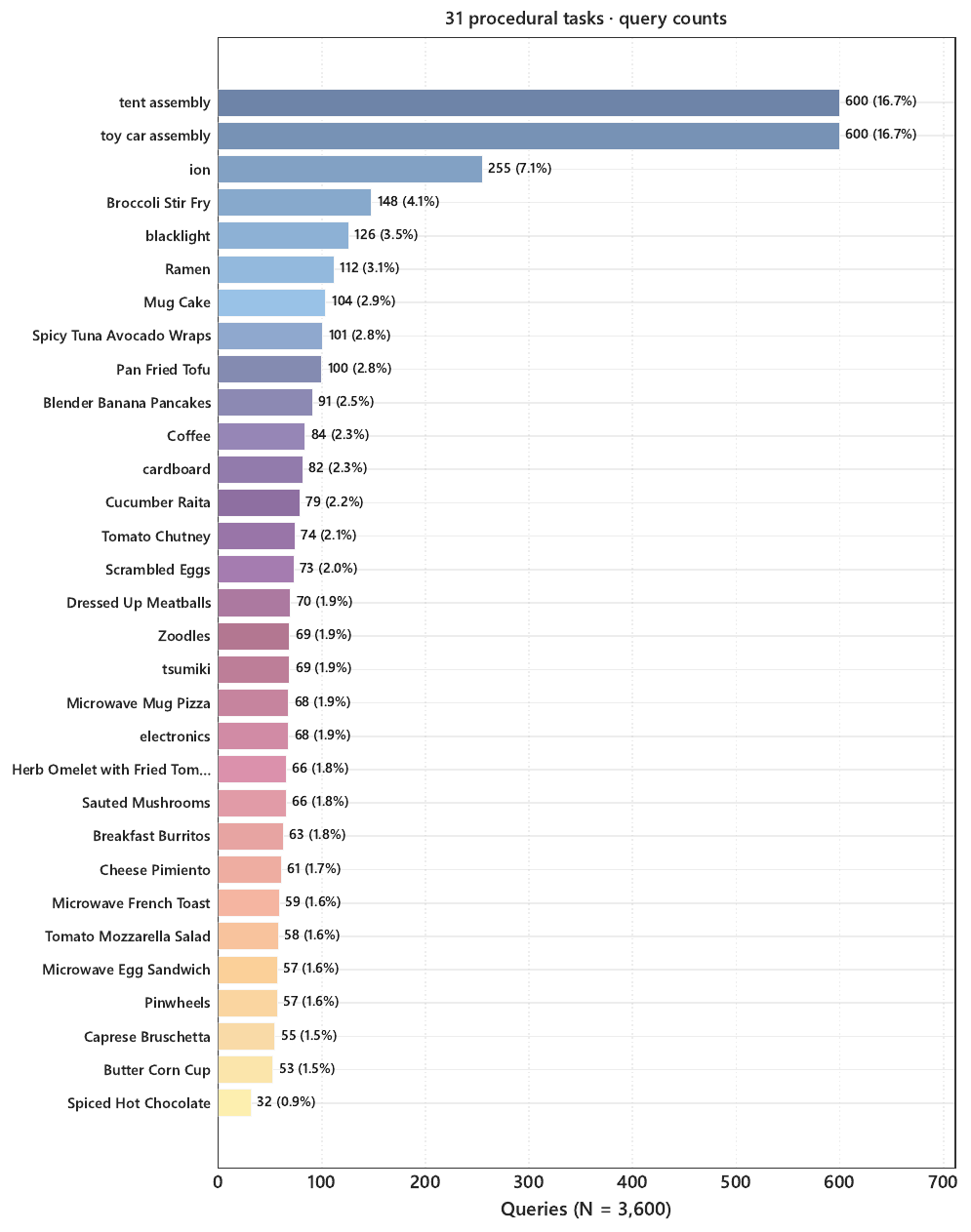}
    \caption{Statistics of 31 procedural task types in EgoProceVQA.}
    \label{fig5}
\end{figure}

\subsection{Implementation Details}

\paragraph{Hardware and software.}
All experiments using local models were conducted using a single NVIDIA GPU with 24~GB VRAM.
For open-source models, 'PyTorch 2.x' with 'bfloat16' precision is used.
To avoid out-of-memory issues during evaluation, GPU memory is released every 20 samples using torch.cuda.empty\_cache().
For experiments using cloud API, Qwen3.5-Plus is parallelised to speed up evaluation using 3,600 samples with up to eight concurrent threads using 'ThreadPoolExecutor'.

\paragraph{Frame sampling.}
For single-clip tasks (Tasks 1, 2, 3, and 6), uniform sampling is used to sample frames from $[s_k, e_k]$.
For multi-clip tasks (Tasks 4 and 5), each clip is sampled independently, where the number of frames per clip is set to $\lfloor N_{\text{total}} / K \rfloor$ with a minimum of 2, and all clips are concatenated into one sequence before feeding them into the models.
Results are reported using two different settings: $N_{\text{total}} \in \{8, 16\}$ to examine sensitivity to temporal resolution.
For API-based models, all frames are resized to 512~px width and encoded as base64 JPEG at quality 85 due to cost constraints.

\paragraph{Answer extraction.}
For multiple-choice tasks, the first occurrence of a standalone letter matching the regular expression \texttt{\textbackslash b([ABCD])\textbackslash b} is extracted from the model output, supporting common response formats such as \texttt{(A)}, \texttt{A.}, \texttt{A:}, \texttt{[A]}, and \textit{''the answer is A''}.
For Task~3, two numeric values are extracted from the model output via regular expressions, corresponding to the predicted start and end times in seconds.

\subsection{Human Performance}

To ensure consistency, a single human annotator independently completed the human performance evaluation. For each of the six task types, 100 questions were randomly sampled to match the scenario proportions used for the models. The annotator watched the full videos and answered the same questions as the models, without applying frame-sampling strategies (e.g., 8 or 16 frames per video), to better reflect natural human video comprehension.

\begin{figure*}
    \centering
    \includegraphics[width=1\linewidth]{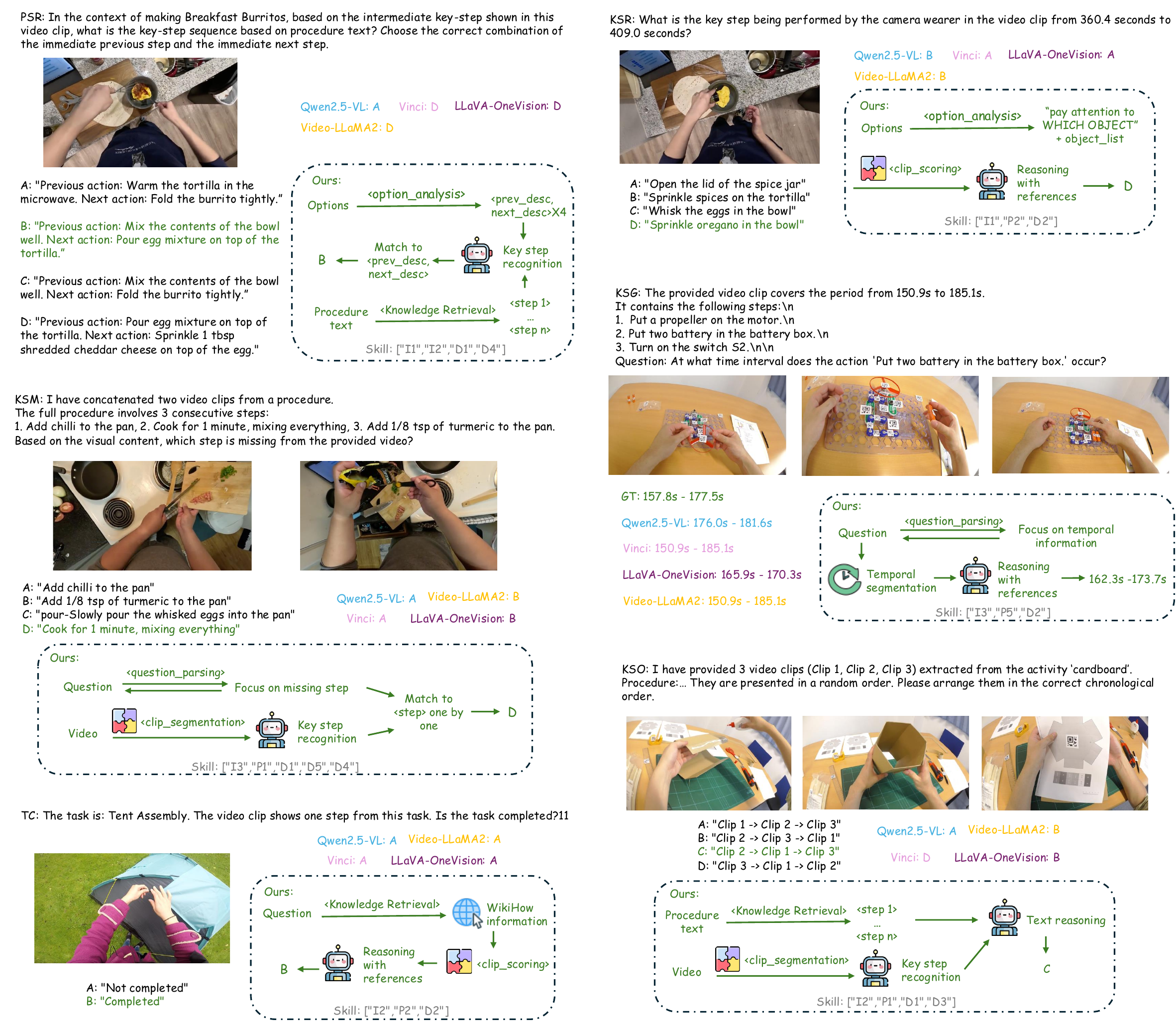}
    \caption{Visual analysis for each type of evaluation task. Here, skills come from the self-exploration outcomes.}
    \label{fig6}
\end{figure*}

\section{Details of Tool Library}
\label{sec4}

\subsection{Grounding DINO: Open-Vocabulary Object Detection}

Grounding DINO enables open-vocabulary object detection conditioned on text queries. Given a video frame $v_t \in \mathcal{V}$ and a set of object hypotheses $\mathcal{C} = \{c_1, c_2, \ldots, c_N\}$ extracted from the question options, Grounding DINO returns a set of detections:
\begin{equation}
    \mathcal{D} = \text{GD}(v_t, \mathcal{C}) = \{(b_j, s_j, l_j)\}_{j=1}^{|\mathcal{D}|},
\end{equation}
where $b_j \in \mathbb{R}^4$ is the bounding box, $s_j \in [0,1]$ is the confidence score, and $l_j \in \mathcal{C}$ is the matched label. Only detections with confidence $s_j \geq \theta_{\text{gd}}$ (we use $\theta_{\text{gd}} = 0.25$) are retained.

To construct the object hypothesis set $\mathcal{C}$, we perform option-driven noun extraction: each option text is tokenized, and a curated stop-word list of verbs, adjectives, and function words is applied to filter out non-physical concepts, retaining only concrete object nouns (eg., ''knife'', ''cucumber'', ''bowl''). This ensures that Grounding DINO searches for visually grounded entities.

\subsection{CLIP: Contrastive Vision-Language Scoring}

CLIP provides a shared embedding space for images and text, enabling zero-shot visual-semantic matching. We employ CLIP in two modes:

\paragraph{Option Scoring.}  Given sampled frames $\mathcal{V}$ and option texts $\{o_k\}_{k=1}^K$, we first extract the average visual feature:
\begin{equation}
    \bar{\mathbf{f}}_v = \frac{1}{T}\sum_{t=1}^{T} \frac{\mathbf{f}_v^{(t)}}{\|\mathbf{f}_v^{(t)}\|_2}, \quad \mathbf{f}_v^{(t)} = \text{CLIP}_{\text{img}}(v_t),
\end{equation}
and the text features $\mathbf{f}_o^{(k)} = \text{CLIP}_{\text{txt}}(o_k)$ for each option. The matching score for option $o_k$ is the cosine similarity:
\begin{equation}
    \text{score}(o_k) = \frac{\bar{\mathbf{f}}_v \cdot \mathbf{f}_o^{(k)}}{\|\bar{\mathbf{f}}_v\|_2 \, \|\mathbf{f}_o^{(k)}\|_2}.
\end{equation}
The resulting similarity ranking is converted to natural language (eg., ''$B(0.27) > A(0.25) > D(0.24) > C(0.23)$'') and injected into the VLM prompt as a visual similarity reference.

\paragraph{Boundary Detection.}  For tasks involving multiple concatenated video clips (eg., clip ordering), we compute the inter-frame cosine similarity curve:
\begin{equation}
    \text{sim}(t) = \frac{\mathbf{f}_v^{(t)} \cdot \mathbf{f}_v^{(t+1)}}{\|\mathbf{f}_v^{(t)}\|_2 \, \|\mathbf{f}_v^{(t+1)}\|_2}, \quad t = 1, \ldots, T-1.
\end{equation}
We apply Gaussian smoothing with $\sigma = 1.0$ to the similarity curve, then identify the two deepest local minima (valleys) as segment boundaries. This exploits the visual discontinuity between clips from different temporal locations.

\subsection{Web Search WikiHow}

To provide procedural prior knowledge for task completion detection, we implement a Web Knowledge Retrieval tool that queries WikiHow\footnote{\url{https://www.wikihow.com}} for structured step-by-step procedure descriptions. Given the activity name $a$ associated with a video, the tool constructs a search query and retrieves the corresponding WikiHow article. The retrieved content is parsed into a structured knowledge base entry:
\begin{equation}
    \mathcal{K}(a) = \{(i, d_i)\}_{i=1}^{N_a},
\end{equation}
where $N_a$ is the total number of steps in the procedure and $d_i$ is the natural language description of the $i$-th step. Of particular importance is the final step description $d_{N_a}$, which characterizes the expected concluding action of the procedure (eg., ''serve the dish'' or ''present the finished product'').

To avoid redundant network requests and ensure reproducibility, all retrieved WikiHow knowledge is cached into a local JSON knowledge base $\mathcal{KB} = \{\mathcal{K}(a)\}$ indexed by normalized activity name. This procedure bank covers all activity types in the evaluation benchmark and is constructed once before inference.

\subsection{Temporal Segmentation}

Video-LLMs receive sparsely sampled frames as input and thus lack the ability to perceive absolute temporal coordinates in the original video. To address thistemporal perception gap, we design a Temporal Segmentation Tool that generates structured time reference anchors.

Given a video clip spanning $[t_s, t_e]$ with $n$ procedural steps as context, the tool uniformly partitions the temporal extent into $n$ equal-duration segments and associates each with its corresponding step description:
\begin{equation}
    \text{TS}(t_s, t_e, n) = \left\{\left(d_i, \, t_s + \frac{(i-1)(t_e - t_s)}{n}, \, t_s + \frac{i(t_e - t_s)}{n}\right)\right\}_{i=1}^{n},
\end{equation}
where $d_i$ is the description of the $i$-th step. The resulting structured reference (eg., ''Step 1 (spread tent): 66.6s -- 86.5s'') is injected into the prompt, providing the model with an approximate temporal reference frame to anchor its reasoning.

Table~\ref{tab:subskills} provides a comprehensive overview of all 12 sub-skills, including their core functions, key outputs, and downstream dependencies.

\begin{table*}[t]
\centering
\caption{Summary of the 12 atomic sub-skills in EgoProceAgent, organized into three layers. Each sub-skill's output is passed via a shared context dictionary to downstream skills. Terminal skills ($^\dagger$) produce the final answer; every valid strategy must end with $D_2$, $D_3$, or $D_4$.}
\label{tab:subskills}
\small
\setlength{\tabcolsep}{4pt}
\renewcommand{\arraystretch}{1.25}
\begin{tabular}{@{}c c l p{5.8cm} p{3.4cm} l@{}}
\toprule
\textbf{Sub-Skill} & \textbf{Layer} & \textbf{Name} & \textbf{Core Function} & \textbf{Key Output(s)} & \textbf{Feeds Into} \\
\midrule
$I_1$ & \multirow{3}{*}{Input $\mathcal{I}$}
  & Option Analysis
  & Identify discriminative dimensions (action, object, quantity) from options; extract noun hypotheses for object detection
  & \texttt{focus\_hint}, \texttt{option\_pairs}, \texttt{object\_hyp.}
  & $P_2, P_3, D_1, D_2$ \\
$I_2$ &
  & Knowledge Retrieval
  & Retrieve procedure steps, and WikiHow knowledge for the video's activity
  & \texttt{procedure\_text}, \texttt{step\_list}, \texttt{wikihow\_steps}
  & $D_1, D_2, D_3, D_4$ \\
$I_3$ &
  & Question Parsing
  & Extract target step name, numbered step list, and clip temporal boundaries from question text and sample metadata
  & \texttt{target\_step}, \texttt{listed\_steps}, \texttt{clip\_start/end}
  & $P_1, P_5, D_1, D_2, D_5$ \\
\midrule
$P_1$ & \multirow{4}{*}{Perception $\mathcal{P}$}
  & CLIP Segmentation
  & Compute inter-frame cosine similarity curve; identify valley minima as clip boundaries; split concatenated video into 2--3 segments
  & \texttt{video\_segments}, \texttt{boundary\_pos.}
  & $D_1, D_2$ \\
$P_2$ &
  & CLIP Scoring
  & Compute frame-averaged visual embedding vs.\ option text cosine similarity; produce a ranked soft-evidence string for the VLM
  & \texttt{clip\_option\_ranking}
  & $D_1, D_2$ \\
$P_3$ &
  & Object Detection
  & Open-vocabulary detection (GroundingDINO) conditioned on option nouns; localize objects in keyframes with confidence scores
  & \texttt{detection\_evidence}
  & $D_2$ \\
$P_5$ &
  & Temporal Reference
  & Uniformly partition clip duration by step count; generate structured time-to-step anchor strings to bridge the VLM temporal perception gap
  & \texttt{temporal\_reference}
  & $D_2$ \\
\midrule
$D_1$ & \multirow{5}{*}{Decision $\mathcal{D}$}
  & Step Identification
  & VLM identifies which procedural step each video segment depicts given a candidate step list; supports single and multi-segment paths
  & \texttt{identified\_steps}, \texttt{identified\_step\_idx}
  & $D_3, D_4, D_5$ \\
$D_2^{\dagger}$ &
  & Evidence Answer
  & VLM answers the question with all accumulated evidence injected into the prompt (focus hint, detections, CLIP ranking, procedure text, temporal reference)
  & \texttt{answer} (letter or interval)
  & --- \\
$D_3^{\dagger}$ &
  & Text Reasoning
  & VLM text-only logical reasoning over identified step indices and procedure order; no video input required
  & \texttt{answer} (letter)
  & --- \\
$D_4^{\dagger}$ &
  & Fuzzy Matching
  & SequenceMatcher maps an identified or missing step description to the best-matching option letter via normalized string similarity
  & \texttt{answer} (letter)
  & --- \\
$D_5$ &
  & Set Elimination
  & Compute set difference: all 3 candidate steps $\setminus$ 2 identified steps $=$ the missing step
  & \texttt{missing\_step\_desc}
  & $D_4$ \\
\bottomrule
\end{tabular}
\end{table*}

\section{Details of EgoProceL Experiment}
\label{sec5}

\textbf{Experiment setting.} In our experiments, to ensure a fair comparison with clustering-based methods evaluated on EgoProceL such as CnC and OPEL, we attempted to keep the overall experimental conditions as consistent as possible. However, Video-LLM is inherently a semantic understanding model, fundamentally different from methods like CnC, making it impossible to construct perfectly identical settings. Therefore, we adopted the following procedure. First, to simulate the crucial ability of key-step recognition and relate it to the semantic understanding capability of Video-LLM, we reformulated key-step recognition as a multiple-choice task. Specifically, we uniformly divided each video into segments corresponding to the number of steps, and required the model to watch each segment and select the key step it belongs to, outputting only the index of the selected step. To further ensure fairness in evaluation, we converted the segment-level predictions of Video-LLM into frame-level labels and then applied the same Hungarian matching algorithm commonly used by these clustering methods, following the evaluation protocol of CnC. Under this setting, our method is in fact at an inherent disadvantage. The uniform segmentation constrains the upper bound of the achievable metrics, and the performance of key-step recognition is directly tied to the model’s semantic understanding. In contrast, methods rely on visual clustering and thus avoid performance degradation caused by semantic understanding errors. Here, the skill strategy result obtained through self-skill-exploration is "I1", "P2", "D2".

\textbf{Experiment results.} As shown in Table~\ref{tab8}, our approach is the only one that is not fine-tuned on any egocentric or procedural understanding tasks. Although RGWOT achieves better performance than ours, our method already demonstrates strong generalization ability in the zero-training setting: we surpass trained methods such as CnC and OPEL. Moreover, under the EgoProceL evaluation setup, the conditions we impose for fair comparison inherently constrain the upper bound of Video-LLM performance on this task; therefore, failing to outperform RGWOT does not imply that our method lacks strong generalization capability.

\begin{table}
    \centering

       \caption{Performances on EgoProceL. Our performance is under 8-frame input.}
    \label{tab8}
\scalebox{0.9}{

    \begin{tabular}{cccccc}
        \toprule
          && \multicolumn{2}{c}{PC Assembly}&\multicolumn{2}{c}{PC Disassembly} \\
         \cmidrule(lr){3-4}\cmidrule(lr){5-6}         
          && F1 & IoU & F1 & IoU\\
         \midrule
       Random   &-& 15.1 & 7.2 & 15.3 & 7.1\\
        Uniform  &-& 17.4 & 8.9 & 18.1 & 9.1\\
       CnC \cite{bansal2022myview}   &trained& 25.1 & 12.8 & 27.0 & 14.8\\
        GPL-2D \cite{bansal2024united}  &trained& 24.0 & 12.6 & 27.4 &15.9 \\
         UG-I3D \cite{bansal2024united}  &trained& 22.0 & 11.7 & 24.2 & 13.8\\
           OPEL \cite{chowdhury2024opel}  &trained& 33.7 & 17.9 & 32.2 & 16.9\\
           RGWOT \cite{mahmood2026procedure}  &trained& 43.6 & 28.0&45.9 & 30.1\\
         \rowcolor{gray!17} Ours &train-free&  39.2& 21.8 & 40.5 & 
19.2\\
           \bottomrule
    \end{tabular}

 }
   
\end{table}

\section{Real-world Demonstration}
\label{sec6}

\begin{figure*}
    \centering
    \includegraphics[width=1\linewidth]{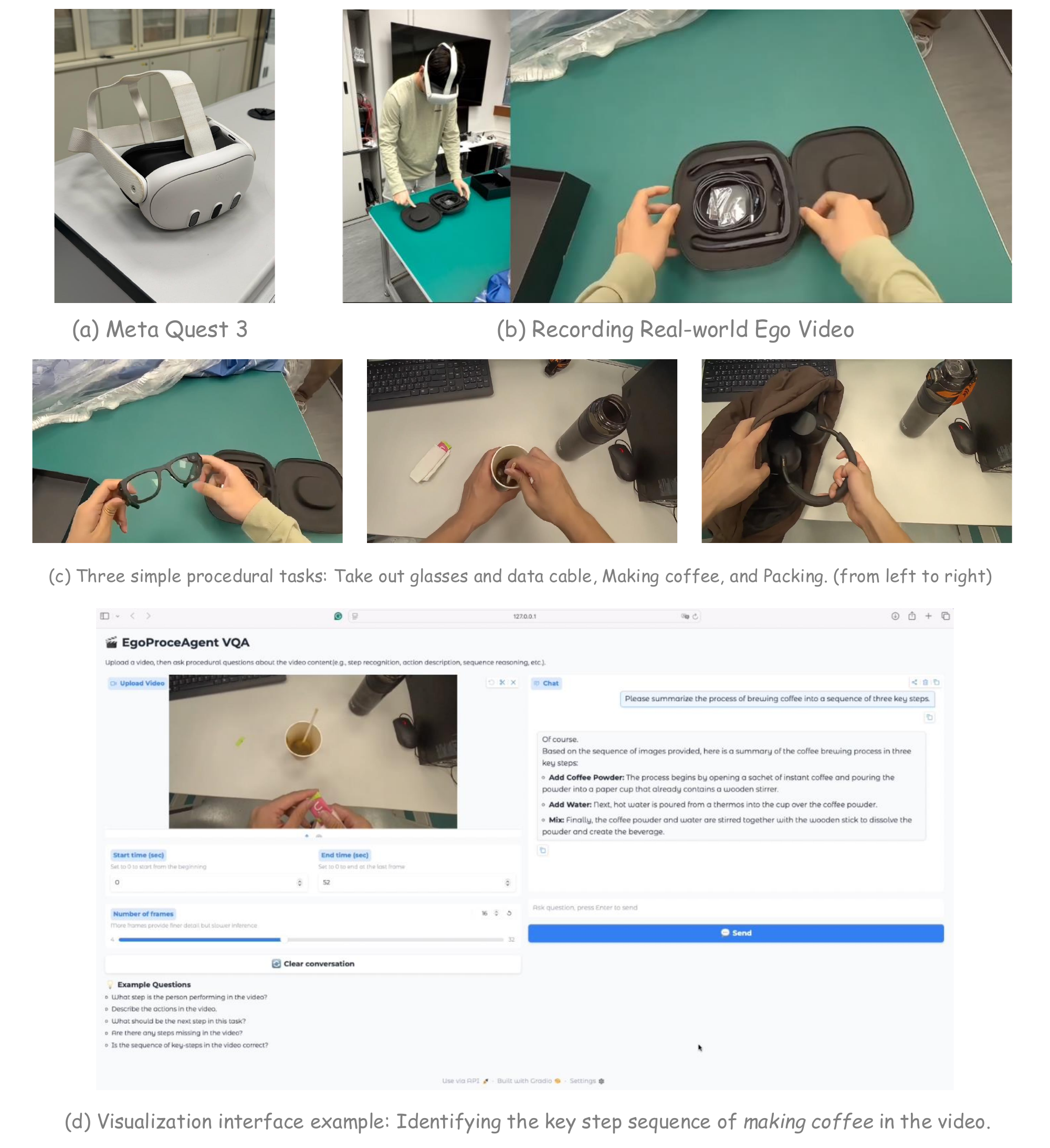}
    \caption{Overview of Real-world Demonstration. (a) shows the data collection device Meta Quest 3. (b) shows the data collection process. (c) shows the three simple procedural tasks we designed. (d) shows our platform built on Gradio.}
    \label{fig7}
\end{figure*}

To present our research results and the immersive experience they provide more intuitively, we recorded a demo video. Specifically, as shown in Fig.~\ref{fig7}, we used \textbf{Meta Quest 3} for egocentric video data collection and have designed three simple yet effective procedural tasks with specific step-by-step constraints: (a) Take out Aria glasses and data cable: 1. "open black box", 2. "open gray bag", 3. "take out Aria", 4. "take out data cable"; (b) Making coffee: 1. "add coffee powder", 2. "add water", 3. "mix them"; (c) Packing: 1. "put the notebook inside", 2. "put the earphone inside", 3. "put the water bottle inside". Although not every step in these tasks exhibits absolute sequential constraints, we provide the procedural step sequence as an instruction manual to EgoProceAgent when needed, requiring the model to use this procedural text as the correct guideline for error detection or next-step action prediction. This meets the requirements for procedural assistance and is therefore effective.

We built our interaction platform using \textbf{Gradio}. Within this platform, users can upload their own videos, configure the desired frame sampling rate, and specify the start and end timestamps of the video segments to be processed. During the dialogue with the model, we demonstrate several forms of assistance for procedural tasks. Specifically, we detect procedural errors characterized by missing steps, perform recognition of key step sequences required to complete a procedural task, and provide auxiliary support that suggests the next action based on the observed video segments and corresponding procedural text. Further implementation details can be seen in the demo video.

\begin{promptbox}[title=System Prompt for Questions without Procedure]
\small
\{question\}\\

Options:\\
A: \{option\_A\}\\
B: \{option\_B\}\\
C: \{option\_C\}\\
D: \{option\_D\}\\
Please answer directly with the option letter (A, B, C, or D).\\

\end{promptbox}

\begin{promptbox}[title=System Prompt for Questions with Procedure]
\small
Procedure for this task:\\
\{proc\_text\}\\

\{question\}\\

Options:\\
A: \{option\_A\}\\
B: \{option\_B\}\\
C: \{option\_C\}\\
D: \{option\_D\}\\
Please answer directly with the option letter (A, B, C, or D).\\

\end{promptbox}

\begin{promptbox}[title= Example for Procedure Text]
\small

"Breakfast Burritos":\\

"To prepare a breakfast burrito, start by whisking an egg in a microwave-safe bowl, then microwave it for 3 minutes, stirring occasionally. Next, mix in 1/2 tbsp sweet and sour sauce, oregano, and 1 tablespoon of salsa, and pour this mixture over a tortilla placed on a cutting board. Top with 1 tbsp shredded cheddar cheese, then roll the tortilla tightly into a log shape, about 1.5 inches thick, ensuring it's sealed properly to hold the filling."\\

\end{promptbox}

\begin{figure*}
\begin{promptbox}[title=Prompt for KSR QA-Pairs Generation]
\small
You are a senior computer vision researcher building an egocentric video comprehension benchmark. 

Your task is to generate highly challenging multiple-choice questions for the "Key Step Recognition" task.

You should strictly output a valid JSON object without any Markdown formatting or extra explanations. The JSON format must be:

\{

"video\_id": "video\_id",

"start\_time": 0.0,

"end\_time": 0.0,

"question": "What is the key step being performed by the camera wearer in the video clip from x seconds to y seconds?",

"options": \{

"A": "Option 1",

"B": "Option 2",

"C": "Option 3",

"D": "Option 4"

\},

"correct\_answer": "A/B/C/D"

\}

I have provided a sequence of keyframes (1 frame per second) extracted from a video clip between \{start\_time:.1f\} seconds and \{end\_time:.1f\} seconds.

The ground truth key step actually performed in this clip is: "\{true\_action\}".

Please carefully observe these frames and generate 3 highly confusing incorrect actions (distractors) based on the visible scene, tools, ingredients, and hand gestures.

[Distractor Generation Strategy] (You must mix the following strategies):

1. Action Confusion (Same Object, Different Action): Keep the target object unchanged but alter the fine-grained hand action (e.g., if GT is "taking tomatoes", distractor could be "washing tomatoes" or "cutting tomatoes").

2. Object Confusion (Same Action, Different Object): Keep the verb unchanged but replace the object with another visible item in the background (e.g., if GT is "stirring with a spoon" and a knife is visible, distractor could be "cutting with a knife").

3. Temporal Confusion: Propose a plausible subsequent or preceding action that might happen immediately before or after this step (e.g., if GT is "pouring oil into pan", distractor could be "turning on the stove").

[Strict Constraints]:

1. NEVER generate absurd distractors involving objects that do not exist in the current visual context.

2. The word count, grammatical structure, and language style of all 4 options (1 Ground Truth + 3 Distractors) MUST BE highly consistent to prevent statistical shortcuts.

3. Randomly shuffle the exact ground truth "\{true\_action\}" with your 3 generated distractors to form options A, B, C, and D.

4. Output ONLY the raw JSON.

\end{promptbox}
\end{figure*}
\begin{figure*}
\begin{promptbox}[title=Prompt for PSR QA-Pairs Generation]
\small

You are a senior computer vision researcher building an egocentric procedural video benchmark for a top-tier conference. 

Your task is to generate highly challenging multiple-choice questions for the "Procedural Sequence Reasoning in Procedure" task.

The goal is to test a vision model's ability to understand long-horizon workflows. By observing ONLY a short intermediate video clip, the model must infer what action logically happened immediately BEFORE, and what action should logically happen immediately AFTER.

You must strictly output a valid JSON object without any Markdown formatting or extra explanations. The JSON format must be:

\{

"video\_id": "video\_id",

"activity\_name": "Activity Task Name",

"anchor\_start\_time": 0.0,

"anchor\_end\_time": 0.0,

"question": "Based on the intermediate step shown in this video clip, what is the most logical procedural sequence? Choose the correct combination of the immediate previous action and the immediate next action.",

"options": \{

"A": "Previous action: [action]. Next action: [action].",

"B": "Previous action: [action]. Next action: [action].",

"C": "Previous action: [action]. Next action: [action].",

"D": "Previous action: [action]. Next action: [action]."

\},

"correct\_answer": "A/B/C/D"

\}

[Procedural Task Context]: 

This clip belongs to a larger procedural task: "\{activity\_name\}".

I have provided a sequence of keyframes (1 frame per second) extracted from an ongoing video clip between \{anchor\_start:.1f\}s and \{anchor\_end:.1f\}s.

The ground truth action happening IN THIS VISIBLE CLIP is: "\{anchor\_desc\}".

According to the ground truth annotation, the true procedural timeline is:

- Previous Action (Just finished before this clip): "\{prev\_desc\}"

- Next Action (Will happen immediately after this clip): "\{next\_desc\}"

Please observe the visual states in the provided frames and generate 3 highly confusing incorrect options (distractors). 

[Distractor Generation Strategy for Temporal Workflow]:

Since this task is specifically "\{activity\_name\}", ALL generated distractors MUST be plausible actions related to making a "\{activity\_name\}". Do not invent absurd actions from other recipes.

1. Sequence Disruption: Use actions that actually belong to the recipe of "\{activity\_name\}" but place them in the wrong chronological order (e.g., placing the "eat sandwich" or "add lettuce" step too early).

2. Logical Reversal: Swap the Previous and Next actions of the ground truth.

3. Plausible but Incorrect Variations: Based on the visible state of the tools/ingredients, suggest actions that a novice might mistakenly do next (e.g., if the egg was just poured, maybe a wrong next action is "stir the egg" instead of "microwave").

[Strict Constraints]:

1. The Ground Truth option MUST exactly be: "Previous action: \{prev\_desc\}. Next action: \{next\_desc\}."

2. The format of all options MUST strictly follow "Previous action: [action]. Next action: [action]."

3. Randomly shuffle the exact Ground Truth with your 3 generated distractors to form options A, B, C, and D.

4. Output ONLY the raw JSON.

\end{promptbox}
\end{figure*}

%\bibliographystyle{ACM-Reference-Format}
%\bibliography{references}

%\end{document}

\end{document}